\documentclass[10pt,twocolumn,letterpaper]{article}

\usepackage{iccv}
\usepackage{times}
\usepackage{epsfig}
\usepackage{graphicx}
\usepackage{amsmath}
\usepackage{amssymb}
\usepackage{ifthen}
\usepackage{times}
\usepackage{epsfig}
\usepackage{graphicx}
\usepackage{color}
\usepackage{microtype}
\usepackage{algorithm}
\usepackage{algorithmic}
\usepackage{ifthen}
\usepackage{booktabs}
\usepackage{enumitem}
\usepackage{caption} 
\newcommand{\final}{0}

\def\etal{\emph{et al}.}

\definecolor{SithColor}{rgb}{0.7,0,0} 
\definecolor{SimingColor}{rgb}{0,0.7,0}
\definecolor{HaitaoColor}{rgb}{0,0,0.7}
\newcommand{\chongyang}[1]{{\color{SithColor} Chongyang: #1}}
\newcommand{\siming}[1]{{\color{SimingColor} Siming: #1}}
\newcommand{\haitao}[1]{{\color{HaitaoColor} Haitao: #1}}

\newcommand{\warning}[1]{{\it\color{red} #1}}
\newcommand{\note}[1]{{\it\color{blue} #1}}
\newcommand{\nothing}[1]{}

\ifthenelse{\equal{\final}{1}}
{
\renewcommand{\chongyang}[1]{}
\renewcommand{\siming}[1]{}
\renewcommand{\haitao}[1]{}
\renewcommand{\warning}[1]{}
\renewcommand{\note}[1]{}
}
{}

\newcommand{\filename}[1]{\url{#1}}
\newcommand{\foldername}[1]{\url{#1}}

\let \bs = \boldsymbol
\let \set = \mathcal

\newcommand{\edge}{\bs{e}}
\newcommand{\cons}{\textup{c}}
\newcommand{\good}{\textup{g}}

\newcommand{\plane}{\textup{Plane}}
\newcommand{\sphere}{\textup{Sphere}}
\newcommand{\cone}{\textup{Cone}}
\newcommand{\type}{\textup{type}}
\newcommand{\emb}{\textup{emb}}
\newcommand{\param}{\textup{param}}
\newcommand{\pull}{\textup{pull}}
\newcommand{\gt}{\textup{gt}}
\newcommand{\push}{\textup{push}}
\newcommand{\cylinder}{\textup{Cylinder}}
\newcommand{\bsplineo}{\textup{B-spline-Open}}
\newcommand{\bsplinec}{\textup{B-spline-Closed}}

\newcommand{\smooth}{\textup{s}}

\newcommand{\realnum}{\mathbb R}

\usepackage[pagebackref=true,breaklinks=true,letterpaper=true,colorlinks,bookmarks=false]{hyperref}

\iccvfinalcopy 

\def\iccvPaperID{1468} 

\begin{document}

\title{HPNet: Deep Primitive Segmentation Using Hybrid Representations}

\author{Siming Yan\textsuperscript{1} \hspace{0.2in}
Zhenpei Yang\textsuperscript{1} \hspace{0.2in}
Chongyang Ma\textsuperscript{2} \hspace{0.2in}
Haibin Huang\textsuperscript{2} \hspace{0.2in}
\\
Etienne Vouga\textsuperscript{1} \hspace{0.3in}
Qixing Huang\textsuperscript{1}
\vspace{4pt}
\\
\textsuperscript{1}{The University of Texas at Austin}\hspace{0.3in} \textsuperscript{2}{Kuaishou Technology}
}

\maketitle
\ificcvfinal\thispagestyle{empty}\fi

\begin{abstract}
This paper introduces HPNet, a novel deep-learning approach for segmenting a 3D shape represented as a point cloud into primitive patches. The key to deep primitive segmentation is learning a feature representation that can separate points of different primitives. Unlike utilizing a single feature representation, HPNet leverages hybrid representations that combine one learned semantic descriptor, two spectral descriptors derived from predicted geometric parameters, as well as an adjacency matrix that encodes sharp edges. Moreover, instead of merely concatenating the descriptors, HPNet optimally combines hybrid representations by learning combination weights. This weighting module builds on the entropy of input features. The output primitive segmentation is obtained from a mean-shift clustering module. Experimental results on benchmark datasets ANSI and ABCParts show that HPNet leads to significant performance gains from baseline approaches.
\end{abstract}

\section{Introduction}
\label{Section:Introduction}

The geometry of man-made objects can be frequently analysed in terms of primitive surface patches (planes, spheres, cylinders, cones, and other simple parametric surfaces). Decomposing a 3D model into primitive surfaces is of fundamental importance with applications in reverse engineering, shape compression, shape understanding, shape editing, and robot learning. \emph{Primitive segmentation} is the task of grouping and labeling points on an object based on primitive shape, and is fundamentally challenging due to the large search space and the fact that primitive patches may only approximately fit the object.

\begin{figure}
\centering
\includegraphics[width=0.48\textwidth]{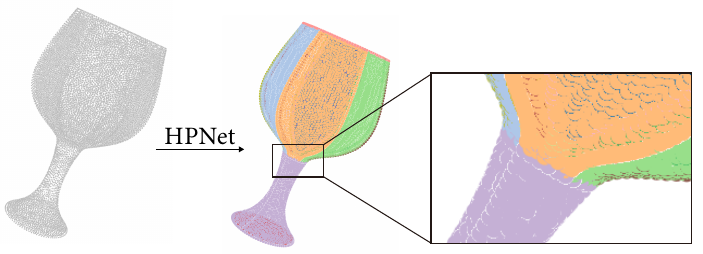}
\caption{HPNet takes a point cloud as input and outputs detected primitive patches. It can handle diverse primitives at different scales. The detected primitives have smooth boundaries.}
\label{Figure:Teaser}
\vspace{-0.1in}
\end{figure}

This paper introduces a deep learning model called HPNet. It takes as input a point cloud (optionally including normals) and outputs a segmentation of the point cloud into primitives, with a type label for each primitive segment (see Figure~\ref{Figure:Teaser}). The main idea of HPNet is to combine traditional tried-and-true geometric heuristics for primitive detection (for instance, algebraic relations between points and shape primitives, and segmentation from sharp edges) with a deep primitive detection approach based on powerful feature learning. We achieve this union through the use of a hybrid point descriptor that combines a learned semantic descriptor and two spectral descriptors. The first spectral descriptor is derived from the adjacency matrix between the input points and predicted primitive parameters. The second spectral descriptor is built based on the adjacency matrix that models sharp edges. In both cases, the spectral approach unifies all primitive segmentation cues as point descriptors. It also rectifies the inputs to the spectral modules, e.g., incorrect predictions of primitive parameters. Given the point descriptors, HPNet employs a mean-shift clustering module to perform primitive segmentation. We also present an effective approach to train HPNet.

A key insight that allows HPNet to succeed on a wide diversity of input shapes is that different types of features have different power on different types of models. For example, when planar shapes are prominent, spectral descriptors are very effective, as eigenvectors of edge-aware adjacency operators localize strongly on the planar patches and segment them robustly.
In contrast, the semantic descriptors are more useful for complex curved shapes, such as patches of cones and cylinders, in which local curvature information is sufficient to determine the shape parameters. HPNet exploits this insight by using adaptive weights to combine different types of features. A weighting module automatically computes the relative weighting for different types of descriptors, based on an entropy measure of each descriptor which estimates the importance of that descriptor for clustering.  


We evaluate HPNet on two benchmark datasets, i.e., ANSI~\cite{li2019supervised} and ABCParts~\cite{Sharma_2018_CVPR}. With only points as input, HPNet improves the Segmentation Mean IoU (MIoU) score from 80.9/75.0 to 91.3/78.1 on the ANSI and ABCParts benchmarks, respectively. With both positions and normals as input, HPNet improves the MIoU score from 88.6/82.1 to 94.2/85.2. We also perform an ablation study to justify the effectiveness of different components of HPNet.
Our code is available at \href{https://github.com/SimingYan/HPNet}{https://github.com/SimingYan/HPNet}.

\section{Related Work}
\label{Section:Related:Works}

Primitive segmentation from 3D shapes has been studied considerably in the past. It is beyond the scope of this paper to provide a comprehensive overview. We refer to~\cite{KaiserZB19} for a thorough survey.  

\paragraph{Non-deep learning based approaches.}
Traditional approaches leverage stochastic paradigms~\cite{fischler1981random,schnabel-2007-efficient,MOD3GP}, parameter spaces~\cite{RABBANI2007355} or clustering and segmentation techniques~\cite{yan2012variational,lafarge:hal-00759265,rtpsrd}. Among these early works, RANSAC~\cite{fischler1981random} and its variants~\cite{schnabel-2007-efficient,Li:2011:GF,MATAS2004837,10.1371/journal.pone.0117341} are the most widely used methods for primitive segmentation and fitting. RANSAC-based methods can estimate model parameters iteratively and showed state-of-the-art results. However, they also suffer from laborious parameter tuning for each primitive. They also do not fully utilize each shape's information (e.g., sharp edges) and prior knowledge about primitive segments, which can be incorporated via machine learning approaches.  

\paragraph{Deep learning based approaches.}
Several recent works~\cite{Sharma_2018_CVPR,abstractionTulsiani17,zou20173d, li2019supervised, SharmaLMKCM20} have studied how to develop deep learning models for primitive segmentation. \cite{zou20173d} and \cite{abstractionTulsiani17} proposed to detect cuboids as rough abstractions of the input shapes. However, their performance is limited on other types of primitives. CSGNet~\cite{sharma2018csgnet} can handle more variety of primitives, but it requires a labeled hierarchical structure for the underlying primitives. This hierarchical structure is not always well-defined.  

Our work is most relevant to SPFN~\cite{li2019supervised} and ParseNet~\cite{Sharma_2018_CVPR}. SPFN proposed a supervised primitive fitting method to predict per-point properties, including segmentation labels, type labels, and normals. Then they introduced a differential model estimation module to fit the primitive parameters. ParseNet proposed a complete model that can handle more primitives, including B-spline patches. However, in the segmentation part of their work, they mainly utilized semantic supervision, which ignored the importance of combining geometric features such as sharp edges. Another novelty of HPNet is that we design the prediction of per-point shape parameters and leverage spectral embedding to generate clearer dense point-wise descriptors.

\paragraph{Hybrid representations for 3D recognition.}
Our approach is motivated by recent methods that use hybrid geometric representations for solving 3D vision tasks. Examples include leveraging different types of keypoints for relative pose estimation~\cite{DBLP:conf/nips/GuibasHL19,yang2020extreme}, utilizing hybrid geometric primitives for 3D object detection and segmentation~\cite{DBLP:conf/cvpr/ZhangLWZH19,ZhangSYH20}, and geometric synthesis under hybrid representations~\cite{DBLP:journals/tog/ZhangYMLHVH20,PoursaeedFAK20}. Our work differentiates from prior methods by learning two spectral descriptors and a weighting sub-module that combines different geometric representations. In particular, the weighting sub-module models entropy for feature representations. This functionality is hard to achieve using alternative techniques, e.g., feature transformation networks. The weighting sub-module is also relevant to the line of work on feature selection~\cite{TangAL14,Liu:2007:CMFS,Li:2017:FS}. The resulting weights are derived from solving a quadratic program.   

\begin{figure*}
\includegraphics[width=\textwidth]{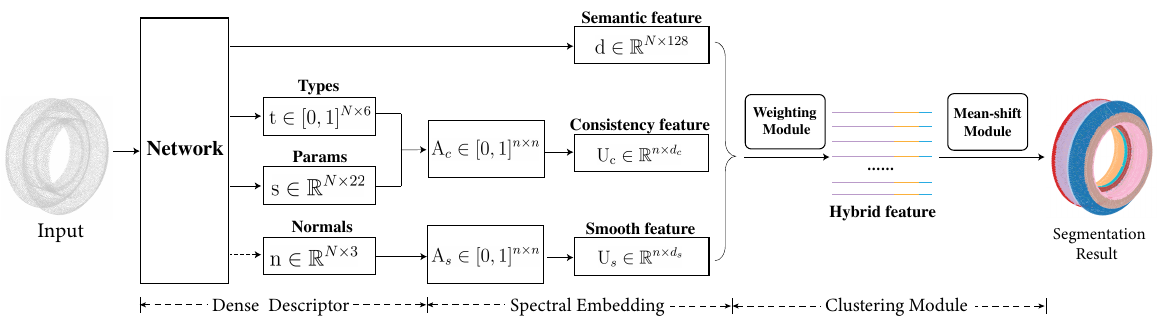}
\caption{Overview of our approach pipeline. HPNet consists of three modules: (1) Dense Descriptor takes a point cloud and optional normal as input and outputs a semantic feature descriptor, a type indicator vector, and a shape parameter vector. (2) Spectral Embedding Module takes dense descriptors as input and builds geometric consistency matrix $A_c$ and smoothness matrix $A_s$. Then it outputs consistency feature $U_c$ and smoothness feature $U_s$. (3) Clustering Module combines three features with adaptive weights and use mean-shift clustering to output the segmentation result.}
\label{Figure:Overview}
\vspace{-0.1in}
\end{figure*}


\section{Overview}
\label{Section:PS:AO}


\subsection{Problem Statement}
\label{Subsec:PS:AO}

We assume the input is given by a point cloud $\set{P} = \{p_i|1\leq i \leq n\}$. Each point $p_i$ has a position $\bs{p}_i\in \realnum^3$ and an optional normal $\bs{n}_i\in \realnum^3$. Our goal is to decompose $\set{P} = \set{P}_1\cup\cdots \cup \set{P}_{K}$ into $K$ primitives $\set{P}_k,1\leq k \leq K$. Each primitive has a type (i.e., plane, sphere, cylinder, cone, b-spline) and some type-specific shape parameters (normal; radius; control points; etc). We translate and scale each point cloud so that its mean is at the origin, and the diameter of the point cloud is $1$.




\subsection{Overview of HPNet}
\label{Subsec:Overview:StoPrimFit}

Figure~\ref{Figure:Overview} illustrates the pipeline of HPNet, which combines three types of modules: dense descriptor, spectral embedding, and clustering. 

\paragraph{Dense descriptor module.} The first module is trained to compute dense point-wise features. These features consist of a semantic descriptor trained to differentiate points from different primitives, a binary type vector identifying the primitive type each point likely belongs to, a parameter vector that predicts the shape parameters of the primitive fitting the point's neighborhood, and the point normal (in case normals were not provided in the input). Note that unlike standard approaches that either learn descriptors to separate different primitive points or detect shape parameters (c.f.~\cite{KaiserZB19,li2019supervised,SharmaLMKCM20}), HPNet combines both approaches. Incorporating both types of features promotes HPNet learning both semantic information (encoded in the type vectors) and local shape geometry (encoded in the shape parameters).

\paragraph{Spectral embedding modules.} The spectral embedding module converts relational cues useful for segmentation (algebraic relations between points that indicate consistency in surface shape parameters, or presence of sharp edges) into dense point-wise descriptors. Each type of relational cue is modeled using an adjacency graph with different weights. The resulting point descriptors are given by the leading eigenvectors of this adjacent matrix. We consider two relational cues that are complementary to the semantic point descriptors. The first module identifies sharp edges, which serve as boundaries between many pairs of primitive patches. Its adjacency matrix assigns high weights to neighboring vertices with similar normals. As the Euclidean distance in the spectral embedding space corresponds to the diffusion distance on the adjacency graph, points that fall on different sides of a sharp edge are far from each other in the embedding space, allowing robust identification of patches separated by creases.

The second module models the consistency between the input points and the predicted shape parameters.  It assigns high weight to an edge if one endpoint is consistent with the local geometry predicted by the other endpoint's shape parameters. The advantage of this formulation comes from the stability of leading eigenvectors against matrix perturbations. Even when a significant portion of the predicted primitive parameters is incorrect, the number of edges with wrong weights due to wrong primitive parameters is typically small, so that the correct predictions still yield an adjacency matrix where each primitive patch is a strongly connected sub-graph. The resulting descriptors are usually cleaner than the raw predicted shape parameters from the dense descriptor module. 

\paragraph{Clustering module.} The clustering module aggregates the point descriptors and applies mean-shift clustering to obtain the final primitive decomposition. A key observation is that the desired combination weights vary across different models. Instead of relying on hand-crafted weights, HPNet learns the combination weights by building on an entropy metric that assesses the underlying clustering structures in  high-dimensional point clouds.

\paragraph{Network training.} We present an effective strategy to train HPNet. The main idea, which has proven to be successful for keypoint-based 6D object pose estimation~\cite{PengLHZB19}, is to use a large-scale training set and a small-scale validation set. The training set is used in learning the dense descriptor module. In contrast, the validation set is used in learning the hybrid parameters of the spectral embedding and clustering modules. This approach alleviates over-fitting. 




\section{Our Method}
\label{Section:Approach}

This section introduces our approach in detail. Section~\ref{Subsec:Dense:Descriptor:Module} to Section~\ref{Subsec:Clustering:Module} present the three modules of HPNet. Section~\ref{Subsec:Network:Training} then introduces how to train HPNet in an end-to-end manner.

\subsection{Dense Descriptor Module}
\label{Subsec:Dense:Descriptor:Module}

As shown in Figure~\ref{Figure:Overview}, the first module of HPNet predicts dense point-wise attributes. The attributes associated with each point $p_i\in \set{P}$ includes a semantic feature descriptor $\bs{d}_i\in \realnum^{128}$, a binary type indicator vector $\bs{t}_i\in \{0,1\}^6$, and a shape parameter vector $\bs{s}_i\in \realnum^{22}$. HPNet considers six primitive types: \plane, \sphere, \cylinder, \cone, \bsplineo, and \bsplinec. The type indicator vector satisfies $t_i(j) = 1$ if and only if the index of the underlying primitive type of $p_i$ is $j$. $\bs{s}_i$ collects shape parameters for the \plane, \sphere, \cylinder, and \cone. HPNet uses the pre-trained SplineNet introduced in ~\cite{SharmaLMKCM20} to get the control points of open and closed b-spline patches. The prediction network is derived from DGCNN~\cite{wang2019dynamic} and Point Transformer~\cite{zhao2020point}, and the details are deferred to the supplemental material.

\subsection{Spectral Embedding Modules}
\label{Subsec:Spectral:Embedding:Module}

The spectral embedding modules take adjacency matrices of the input point cloud as inputs and output their leading eigenvectors. Each leading eigenvector is then a dense point descriptor function. HPNet considers two adjacency matrices. The first one models the consistency between the positions of the input points and the predicted shape parameters in a neighborhood around the point. The second one models presence of sharp edges. We provide the details for constructing these matrices in Sections~~\ref{Subsec:Consistency:Mat} and~\ref{Subsec:Smoothness:Mat} below.

\subsubsection{Geometric Consistency Matrix}
\label{Subsec:Consistency:Mat}

We first define a distance metric $d(p_i,\bs{s}_j)$ between one point $p_i$ and the predicted shape parameter $\bs{s}_j$ associated with another point $p_j$~\cite{li2019supervised}. Let $t_j$ be the predicted primitive type of $s_j$. When $t_j\in \{\plane,\sphere,\cone,\cylinder\}$, we define $d(p_i, \bs{s}_j) = $
$$
\left\{
\begin{array}{cl}
|\bs{p}_i^T\bs{n}_j-d_j| & t_j = \plane \\
|\|\bs{p}_i-\bs{o}_j\|-r_j| & t_j = \sphere \\
|\|(I-\bs{a}_j\bs{a}_j^T)(\bs{p}_i-\bs{o}_j)\|-r_j| & t_j = \cylinder \\
{\scriptsize \|\bs{p}_i-\bs{o}_j\|\cos\left(\arccos\left(\bs{a}_j^T\frac{\bs{p}_i-\bs{o}_j}{\|\bs{p}_i-\bs{o}_j\|}\right)-\theta_j\right) } & t_j = \cone
\end{array}
\right.\
$$
where $\bs{n}_j$ and $d_j$ denote the normal and distance of a plane primitive; $\bs{o}_j$ and $r_j$ the center and radius of a sphere primitive; $\bs{a}_j$, $\bs{o}_j$, and $r_j$ the direction, center, and radius of a cylinder; and $\bs{o}_j$, $\bs{a}_j$, and $\theta_j$ the center, apex, and angle of a cone. There are $22$ total parameters.

When $t_j\in \{\bsplineo,\bsplinec\}$, we define $d(p_i, \bs{s}_j)$ as the closest distance between $p_i$ and the B-spline patch specified by $\bs{s}_j$. Please refer to the supplemental material for an efficient approach for computing $d(p_i, \bs{s}_j)$ approximately. 

We then define the signed weight between $p_i$ and $\bs{s}_j$ as
$$
w(p_i,\bs{s}_j):= \exp\left(-\frac{d^2(p_i,\bs{s}_j)}{2\sigma_{t_j}^2}\right),
$$
where $\sigma_{t_j} > 0$ is a hyperparameter associated with type $t_j$. Finally, we define the consistency adjacency matrix $A_{c}\in [0,1]^{n\times n}$, whose elements are given by
$$
A_{\cons}(i,j) = \big(w(p_i,\bs{s}_j)+w(p_j,\bs{s}_i)\big)/2, \quad 1\leq i,j \leq n.
$$
Let $\lambda_{\cons,i}$ and $\bs{u}_{\cons,i}$ denote the $i$-th eigenvalues and eigenvectors of $A_{\cons}$. We define the resulting descriptors as columns of $U_{\cons} = \big(\sqrt{\frac{\lambda_1}{\lambda_1}}\bs{u}_{\cons,1},\cdots, \sqrt{\frac{\lambda_1}{\lambda_{d_{\cons}}}}\bs{u}_{\cons,d_{\cons}}\big)\in \realnum^{n\times d_{\cons}}$. Depending on the number of primitives, $d_{\cons}$ varies across different shapes in our experiments. 

\paragraph{Discussion.}
Below we provide an analysis to show that the spectral descriptor is superior to the predicted primitive parameters. 
Denote $A_{\cons}^{\good}$ as the matrix $A_{\cons}$ with all entries $(i,j)$ set to zero when $p_i,p_j$ belong to different primitives. Without losing generality, we can reorder the vertices so that
$A_{\cons}^{\good} = \textup{diag}(A_{\cons,1}^{\good},\cdots,A_{\cons,K}^{\good})$ is a block diagonal matrix, where $A_{\cons,k}^{\good}$ contains the weights for pairs of points belonging to the $k$-th primitive. Decompose
$$
A_{\cons} = A_{\cons}^{\good} + E.
$$
To simplify the discussion, we further assume the weights are binary, i.e., $1$ for consistent pairs and $0$ for inconsistent pairs. We also assume $d_c = K$ in this analysis for convenience. 

Let $U_{\cons}^{\good}\in R^{n\times K}$ be the counterpart of $U_{\cons}$, whose columns are the re-scaled eigenvectors of $A_{\cons}^{\good}$. A variant of the Davis-Kahan theorem~\cite{yu2014useful} provides the difference between $U_{\cons}^{\good}$ and $U_{\cons}$:
\begin{equation}
\min\limits_{R\in O(K)}\|U_{\cons}R-U_{\cons}^{\good}\|_{\set{F}} \leq \frac{\sqrt{\lambda_{1}(A_{\cons}^{\good}})\|E\|_{\set{F}}}{\lambda_{K}(A_{\cons}^{\good})-\lambda_{K+1}(A_{\cons}^{\good})},
\label{Eq:DK}
\end{equation}
where $\|\cdot\|_{\set{F}}$ denotes the Frobenius norm. 

Applying (\ref{Eq:DK}), we argue that when primitive sizes are comparable, i.e., the size of the $k$-th primitive $n_k = \frac{n}{K}$, 
\begin{equation}
\min\limits_{R\in O(K)}\frac{\|U_{\cons}R-U_{\cons}^{\good}\|_{\set{F}}}{\|U_{\cons}^{\good}\|_{\set{F}}} = O\left({\scriptstyle \frac{2K\sqrt{\rho}}{(1-\rho)+\sqrt{1-\rho}}} n^{-\frac{1}{4}}\right),
\label{Eq:Spectral:Embedding:Error}
\end{equation}
where $\rho$ is the fraction of outliers of the predicted shape parameters. Moreover, it is easy to see that corresponding error of the predicted shape parameters scales as $O(\sqrt{\rho})$. It follows when $n$ is sufficiently large, the spectral descriptor is superior to the predicted shape parameters. Due to space constraint, we leave the proof to the supplemental material.

\vspace{-0.1in}
\subsubsection{Smoothness Matrix}
\label{Subsec:Smoothness:Mat}

Consider a $k$-nearest neighbor graph $\set{G} = (\set{P},\set{E})$ whose vertices are taken from the input point cloud ($k = 50$ in this paper). For each edge $(i,j)\in \set{E}$, we define the corresponding weight as
$$
w_{(p_i,p_{j})} = \exp\left(-\frac{\|\bs{n}_i-\bs{n}_{j}\|^2}{2\sigma_{\edge}^2}\right),
$$
where $\sigma_{\edge}$ is a hyperparameter, and where $\bs{n}_i$ and $\bs{n}_{j}$ are the normals at $p_i$ and $p_{j}$, respectively.
Let $A_{\smooth}$ be the weighted adjacency matrix of $\set{G}$. The spectral descriptors $U_{\smooth}$ associated with $A_{\smooth}$ are then defined identically to $U_{\cons}$ (columns are scaled leading eigenvectors of $A_{\smooth}$).

\vspace{-0.1in}
\paragraph{Discussion.} The usefulness of $U_{\smooth}$ comes from the fact the Euclidean distance between $p_i$ and $p_j$ defined by the rows of $U_{\smooth}$ is identical to the diffusion distance on the weighted graph $\set{G}_{\smooth}$ specified by $A_{\smooth}$ (c.f.~\cite{Sun:2009:HKS}).
Note that points that are close to each other on the input model, but are on different sides of sharp edges, tend to have large diffusion distances (the paths connecting them have to detour around sharp edges). Therefore, these points have large distances in the embedding space.
\begin{table*}
  \footnotesize
  \centering
  \begin{tabular}{l|c|c|c|c|c|c|c|c|c}
  \toprule
  &  & \multicolumn{4}{c|}{ANSI} & \multicolumn{4}{c}{ABCParts}\\
  \hline
  & Input & seg iou & type iou & res error & P coverage & seg iou & type iou & res error & P coverage \\
  \hline
  NN       & p   & 81.92 & 95.00 & 0.014 & 91.90 & 54.10 & 61.10 & - & - \\
  RANSAC~\cite{schnabel2007efficient}  & p+n & 70.10 & 93.13 & 0.029 & 78.79 & 67.21 & -     & 0.022 & 83.40 \\
  \hline
  SPFN~\cite{li2019supervised}     & p   & 77.61 & 95.43 & 0.014 & 92.10 & 58.15 & 73.88 & 0.023 & 87.58 \\
  SPFN~\cite{li2019supervised} & p+n & 88.05 & 98.10 & 0.011 & 92.94 & 73.41 & 80.04 & 0.020 & 89.40 \\
  ParseNet~\cite{SharmaLMKCM20} & p   & 80.91 & 97.49 & 0.013 & 90.91 & 75.01 & 81.16 & 0.014 & 87.95 \\
  ParseNet~\cite{SharmaLMKCM20} & p+n & 88.57 & 98.26 & 0.010 & 92.72 & 82.14  & 88.60 & 0.011 & 92.97 \\  
  \hline
  Ours-wc  & p   & 90.12 & 98.21 & 0.012 & 92.00 & 76.71 & 82.14 & 0.013 & 88.21 \\
  Ours-wc  & p+n & 92.41 & 98.87 & 0.010 & 93.58 & 83.20 & 89.54 & 0.010 & 93.15 \\
  Ours     & p & 91.34 & 98.66 & 0.011 & 93.02 & 78.12 & 85.32 & 0.012 & 90.54  \\
  Ours     & p+n & \textbf{94.15} & \textbf{98.90} & \textbf{0.008} & \textbf{94.02} & \textbf{85.24} & \textbf{91.04} & \textbf{0.009} & \textbf{94.31} \\
  \bottomrule
  \end{tabular}
{%
 \caption{Benchmark evaluation on our approach and baseline approaches. We provide different input to the model:points(p) and points+normals(p+n). Here, Ours-wc stands for our method without combining two spectral descriptors.}
 \label{tab:baseline-comparison}
}
\end{table*}
\subsection{Clustering Module}
\label{Subsec:Clustering:Module}

The clustering module concatenates the different types of point descriptors defined above, namely, the semantic descriptors $\bs{d}_i$ and the two spectral features specified by $U_{\cons}^{T}\bs{e}_i$ and $U_{\smooth}^T\bs{e}_i$. As mentioned in the introduction, the relative importance of the different descriptors varies dramatically based on the object geometry; so rather than using fixed weights to combine the descriptors, we introduce an approach that learns how to combine them.

\paragraph{Weighting sub-module.} To make the notations uncluttered, we describe our weighting scheme in a generic settings where there are $L$ features $F_l\in \realnum^{n\times m_l}, 1\leq l \leq L$, with $m_l$ the dimension of the $l$-th feature. Our goal is to compute a weight $w_l \in (0,1)$ for each feature, with $\sum_l w_l^2 = 1$. In the context of this paper, $L = 1 + d_{\cons} + d_{\smooth}$, i.e., $F_1$ corresponds the semantic descriptors $\bs{d}_i$, and each remaining $F_l$ corresponds to one spectral descriptor. Note that we weight each spectral descriptor individually because the desired number of spectral descriptors is dependent on the number of underlying primitives, which varies across different shapes.

HPNet uses the criterion that a feature $F_l$ should have large weights if $F_l$ reveals articulated cluster structure. Motivated from the feature selection for clustering approach described in~\cite{Dash:2000:FSC}, HPNet applies an entropy score to define the feature weight. Specifically, we first model the multivariate probability function of the feature space $F_l$ as
$$
P_{F_l}^{\sigma_l}(\bs{x}):= \frac{1}{n}(2\pi)^{-\frac{m_l}{2}}\sigma_l^{-m_l}\sum\limits_{i=1}^{n}\exp\left(-\frac{\|\bs{x}-F_l^T\bs{e}_i\|^2}{2\sigma_l^2}\right),
$$
where $\sigma_l$ is a hyperparameter associated with $F_l$. The entropy of $F_l$ is then
\begin{equation}
H_{\sigma_l}(F_l):= -\sum\limits_{i=1}^{n}P_{F_l}^{\sigma_l}(F_l^T\bs{e}_i)\log\Big(P_{F_l}^{\sigma_l}(F_l^T\bs{e}_i)\Big).
\label{Eq:Entropy:Fl}
\end{equation}
Intuitively, features where points form clusters (versus random point distributions) tend to have low entropy values. We model the weight of each feature $F_l$ so that it is inversely proportional to $H_{\sigma_l}(F_l)$:
\begin{equation}
w_l := \overline{w}_l/\sqrt{\sum\limits_{l}\overline{w}_l^2}, \quad \overline{w}_l:= \frac{1}{H_{\sigma_l}(F_l)}.    
\end{equation}


\paragraph{Mean-shift clustering sub-module.}
Since the primitive number varies between different models, we apply a mean-shift clustering procedure~\cite{comaniciu2002mean} to obtain the primitive segmentation result.

\begin{figure*}
\centering
\footnotesize
\def\imh{0.073\textwidth}
\def\imw{0.15\textwidth}
\newcommand{\TT}[1]{\raisebox{-0.5\height}{#1}}
\setlength{\tabcolsep}{1pt}
\begin{tabular}{cccccccc}

\rotatebox[origin=c]{90}{G.T} & 
\TT{\includegraphics[width=\imw]     {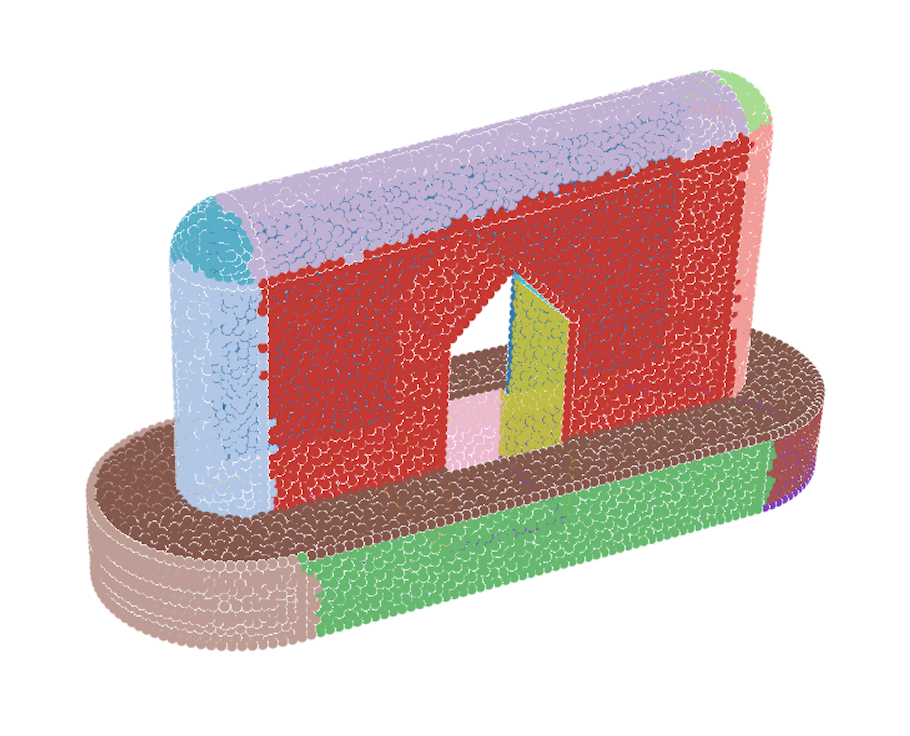}} & 
\TT{\includegraphics[width=\imw]   {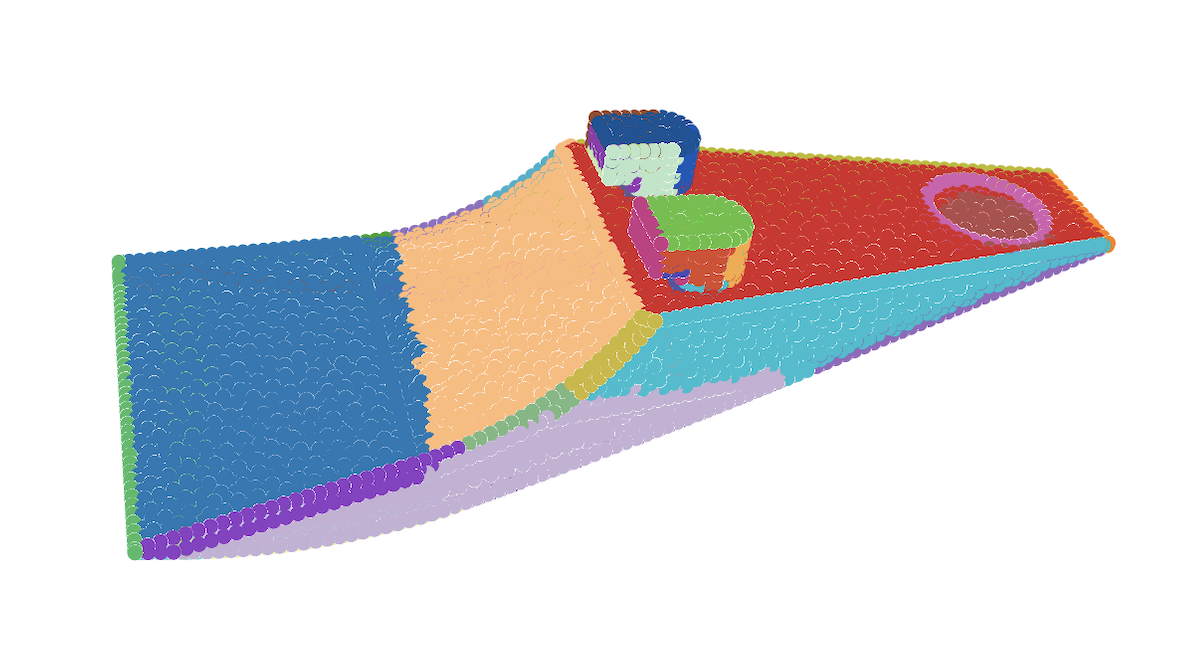}} & 
\TT{\includegraphics[height=0.11\textwidth]      {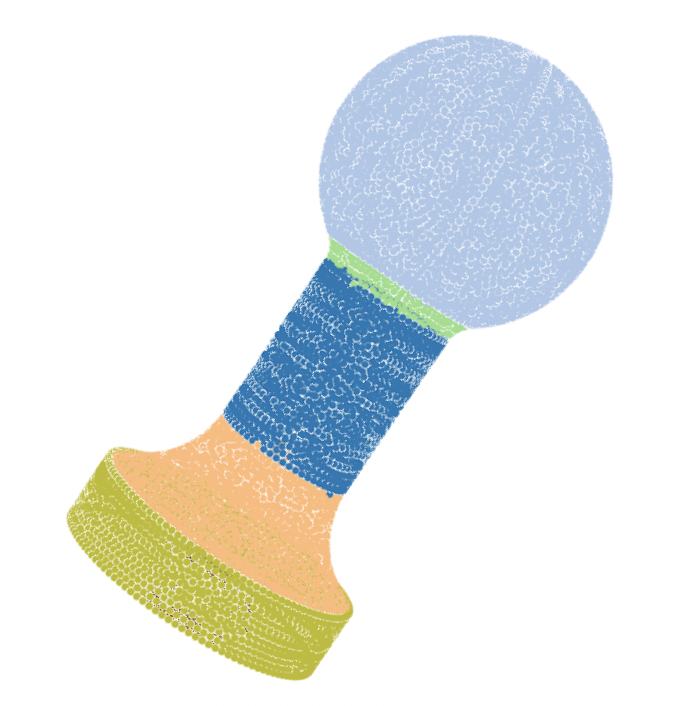}} & 
\TT{\includegraphics[width=\imw]    {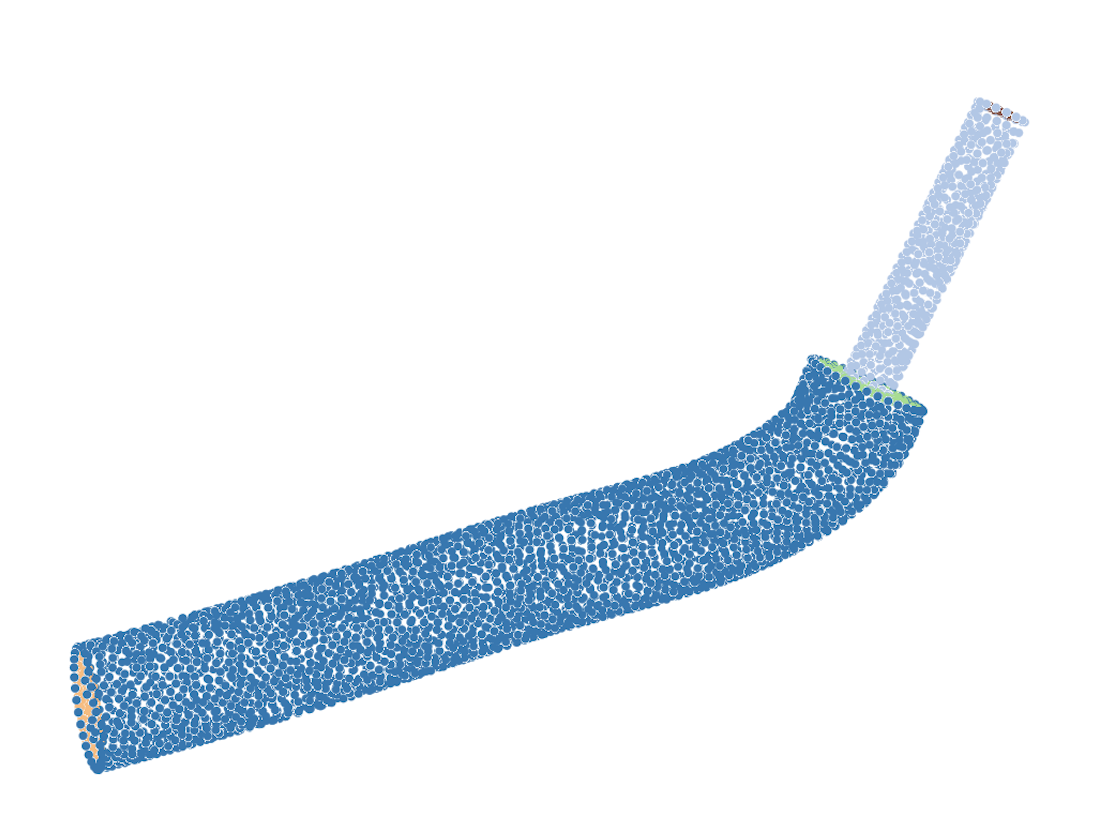}} & 
\TT{\includegraphics[width=0.12\textwidth]    {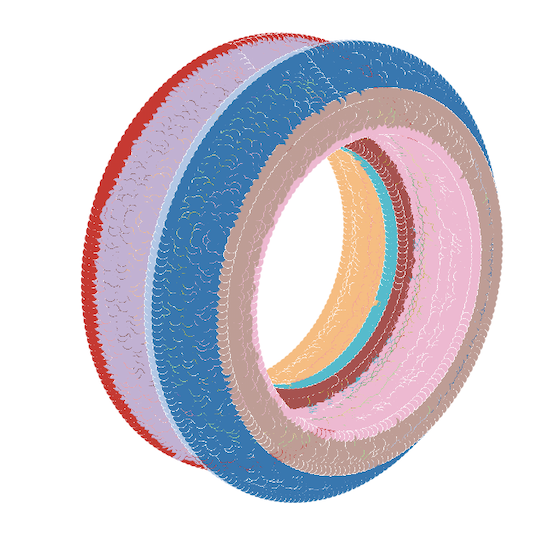}} & 
\TT{\includegraphics[width=0.13\textwidth]      {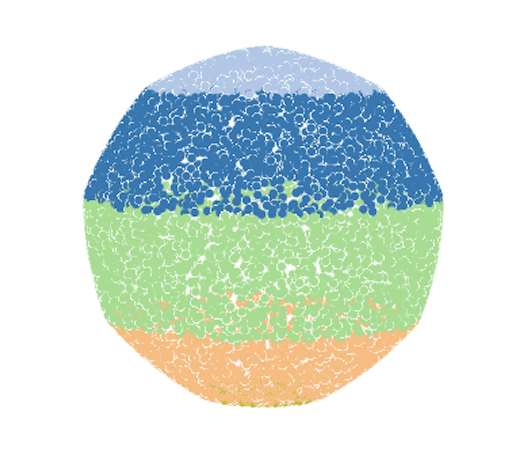}} &
\TT{\includegraphics[width=0.13\textwidth]      {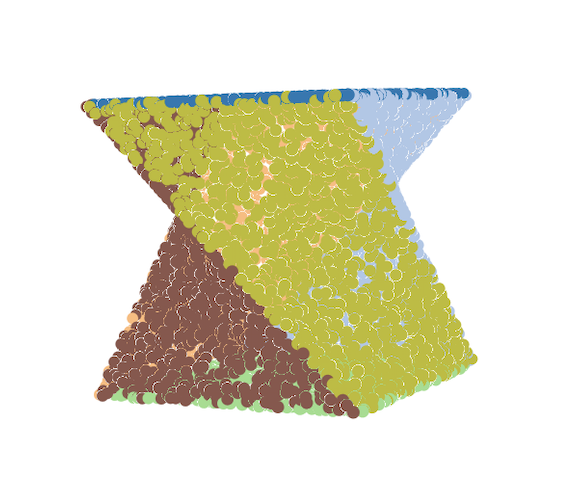}} \\
\hline
\rotatebox[origin=c]{90}{SPFN} & 
\TT{\includegraphics[width=\imw]     {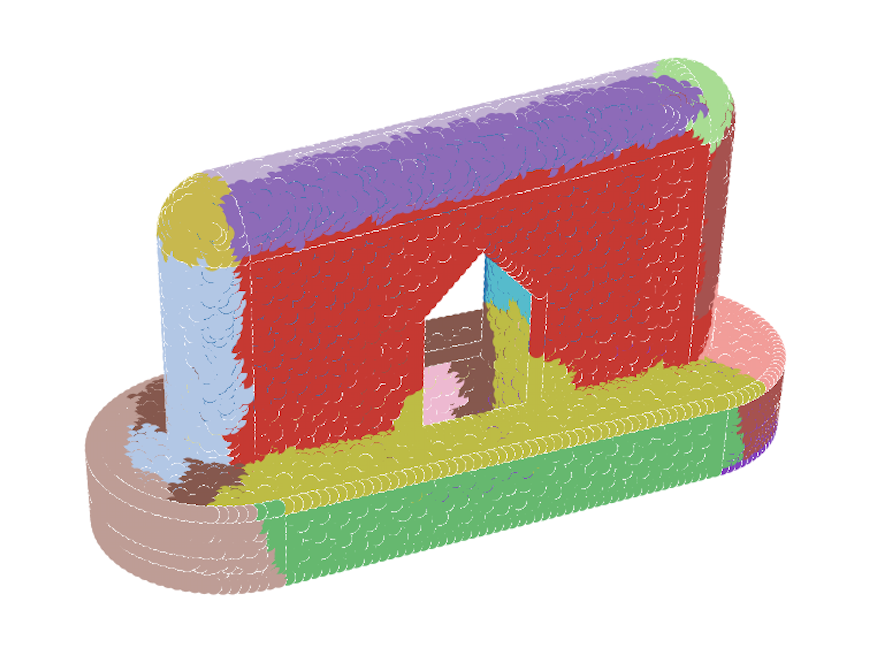}} &
\TT{\includegraphics[width=\imw]   {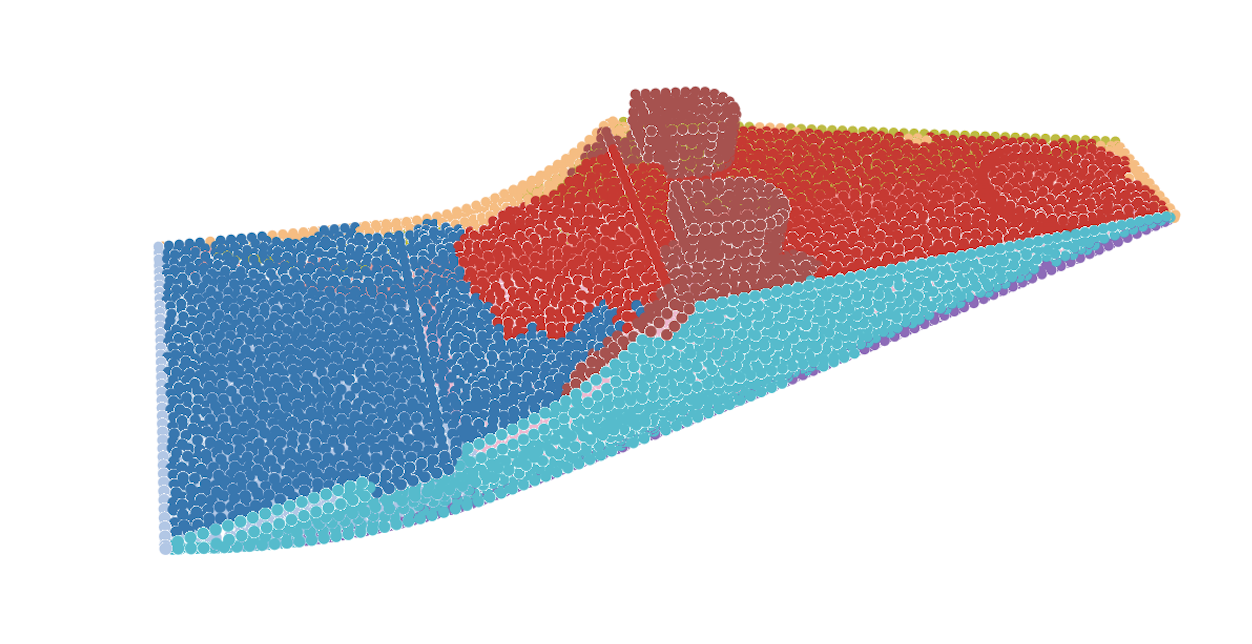}} & 
\TT{\includegraphics[height=0.11\textwidth]      {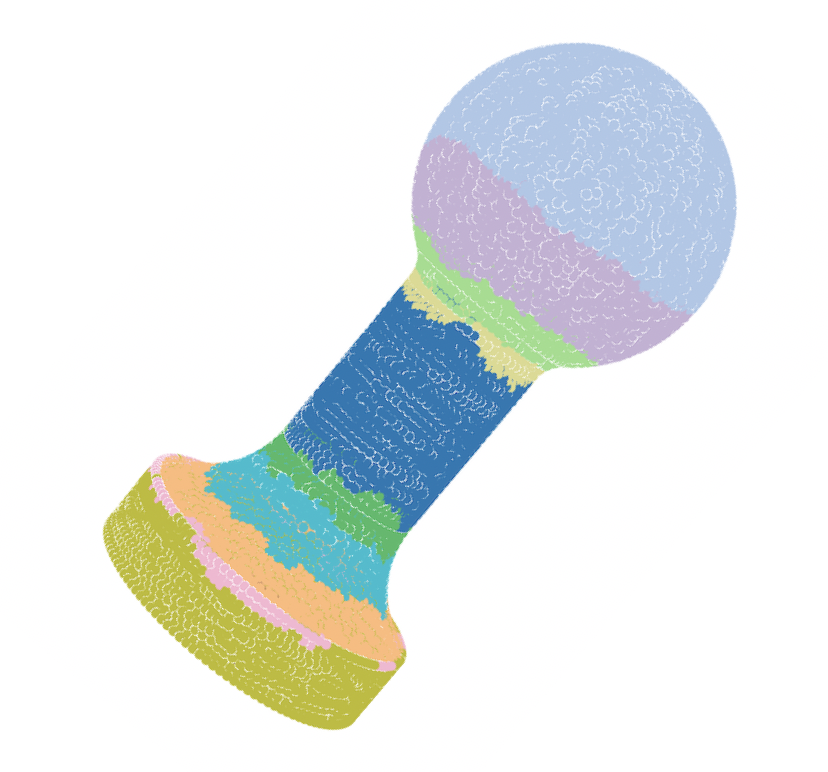}} & 
\TT{\includegraphics[width=\imw]    {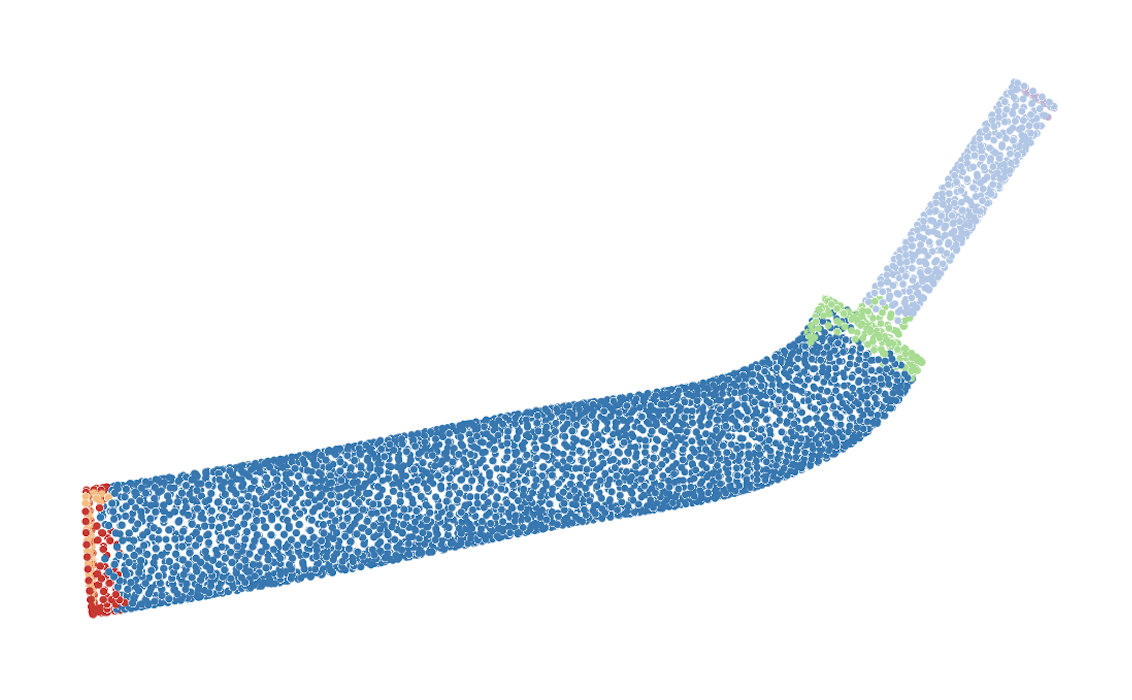}} & 
\TT{\includegraphics[width=0.115\textwidth]    {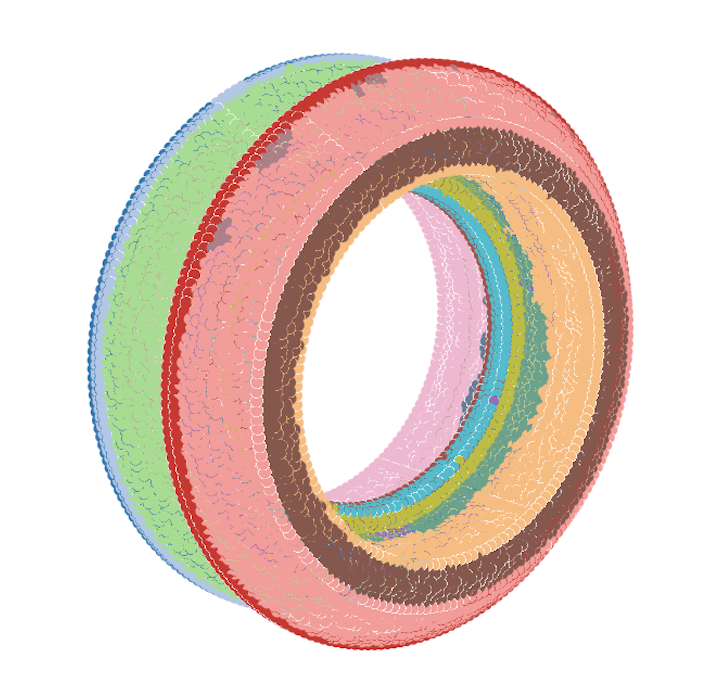}} &
\TT{\includegraphics[width=0.13\textwidth]      {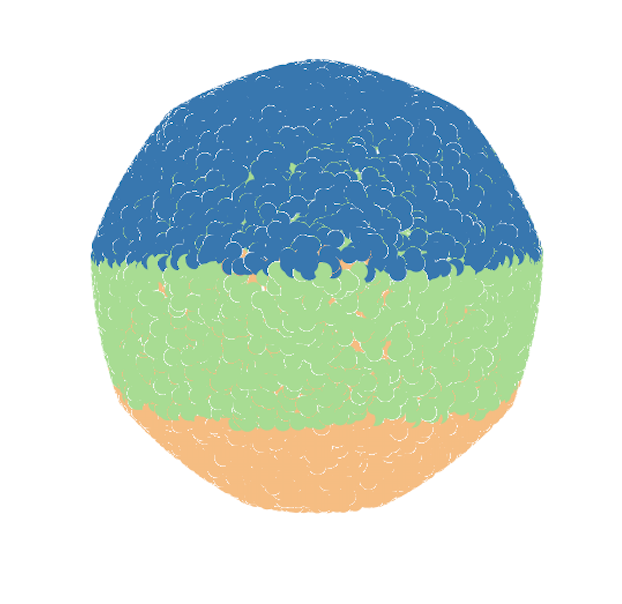}} &
\TT{\includegraphics[width=0.13\textwidth]      {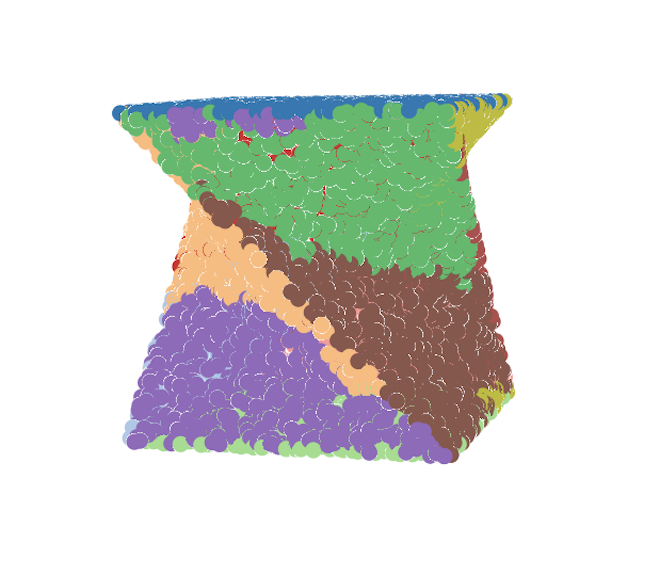}}\\

\rotatebox[origin=c]{90}{Parsenet} & 
\TT{\includegraphics[width=\imw]     {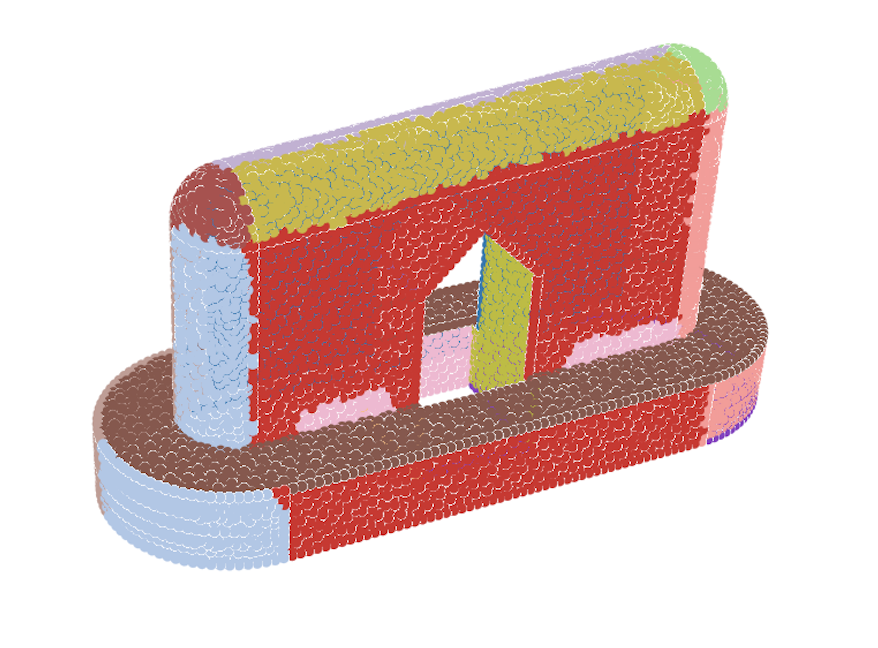}} & 
\TT{\includegraphics[width=\imw]   {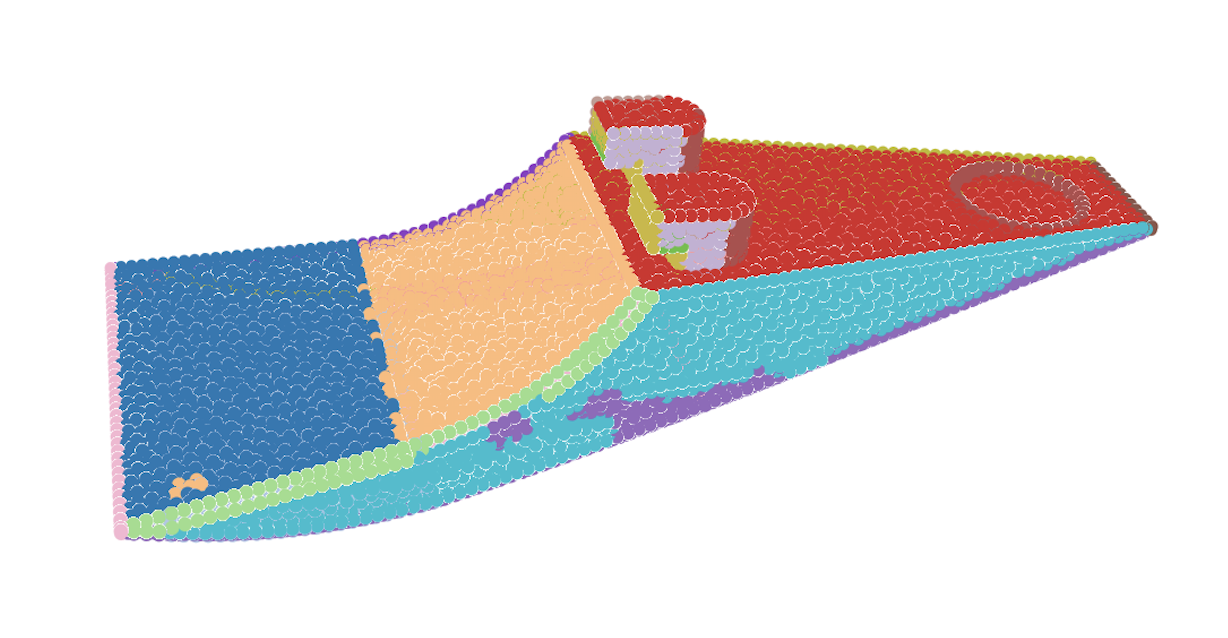}} & 
\TT{\includegraphics[height=0.11\textwidth]      {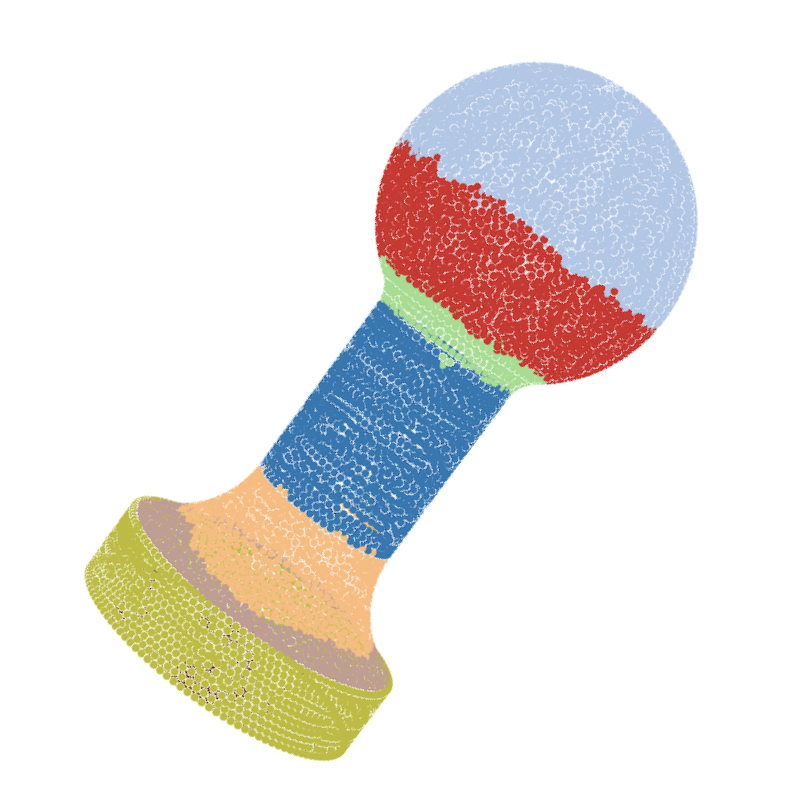}} & 
\TT{\includegraphics[width=\imw]    {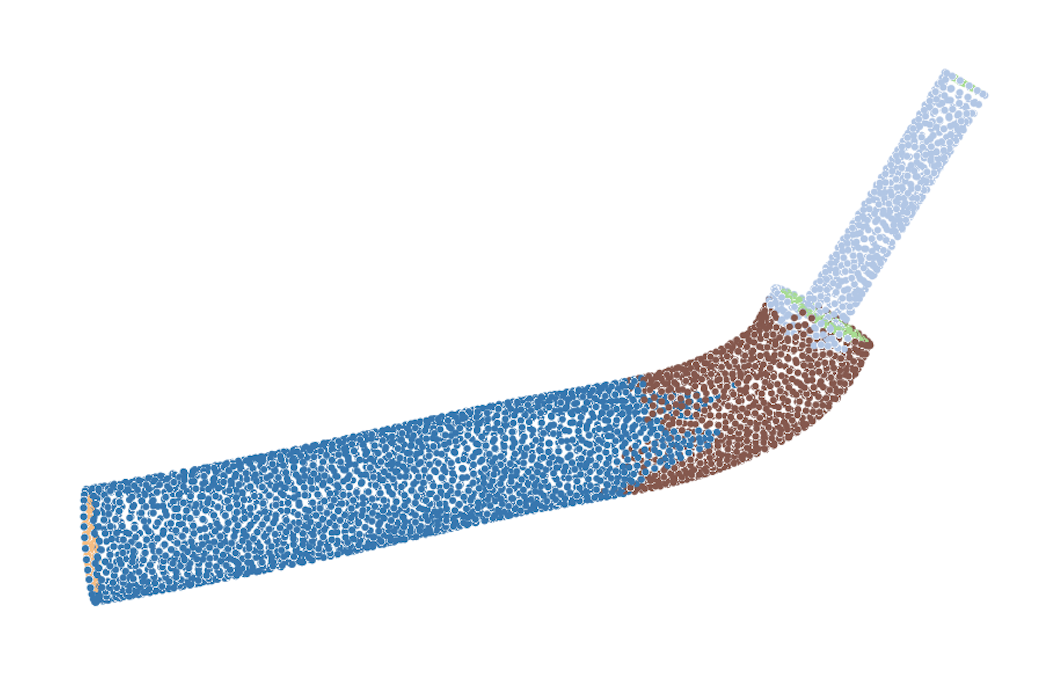}} & 
\TT{\includegraphics[width=0.125\textwidth]    {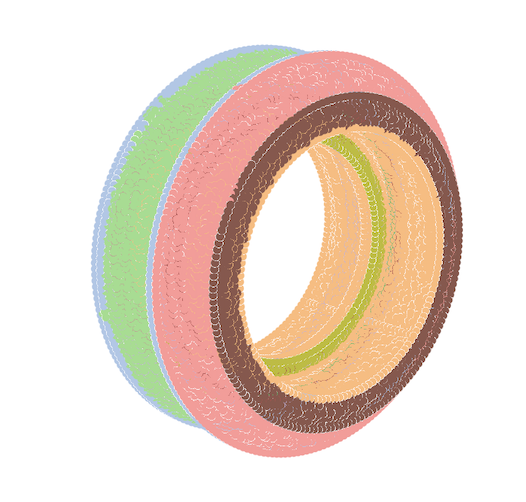}} &
\TT{\includegraphics[width=0.13\textwidth]      {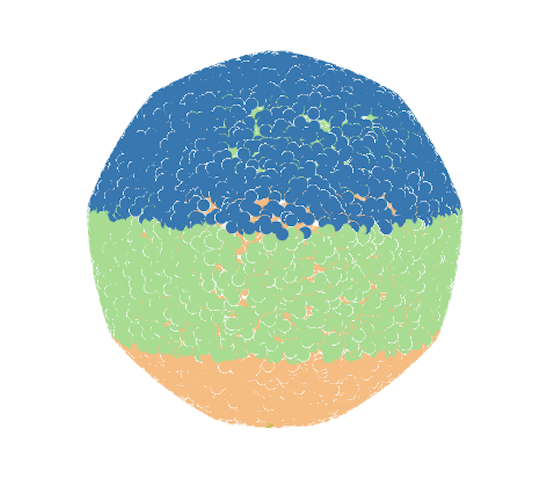}} &
\TT{\includegraphics[width=0.13\textwidth]      {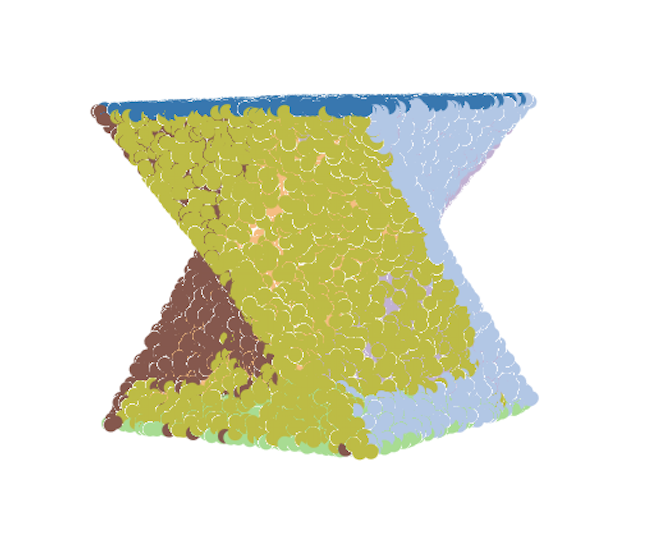}}\\

\rotatebox[origin=c]{90}{Ours} & 
\TT{\includegraphics[width=\imw]     {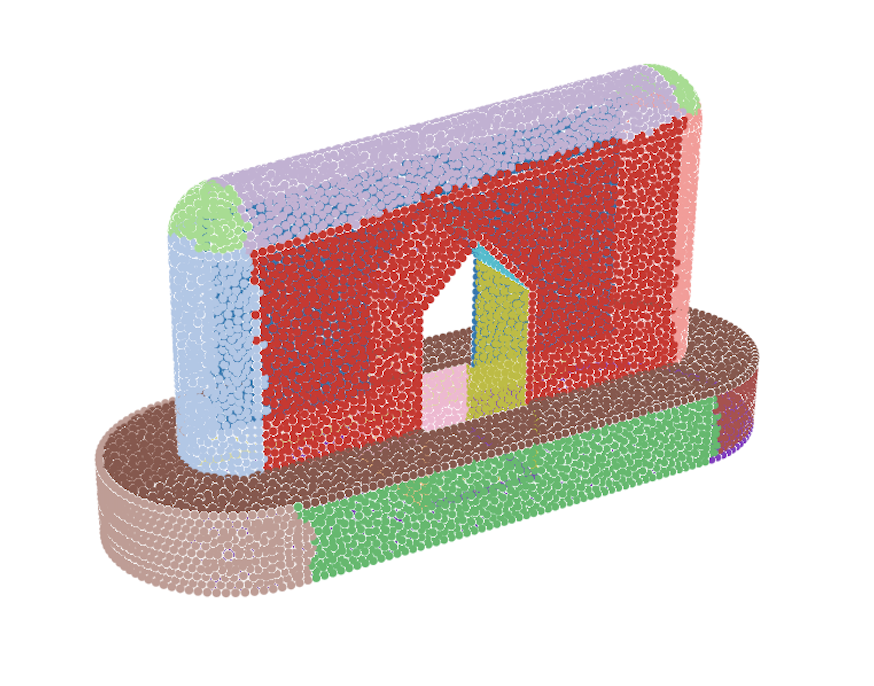}} & 
\TT{\includegraphics[width=\imw]   {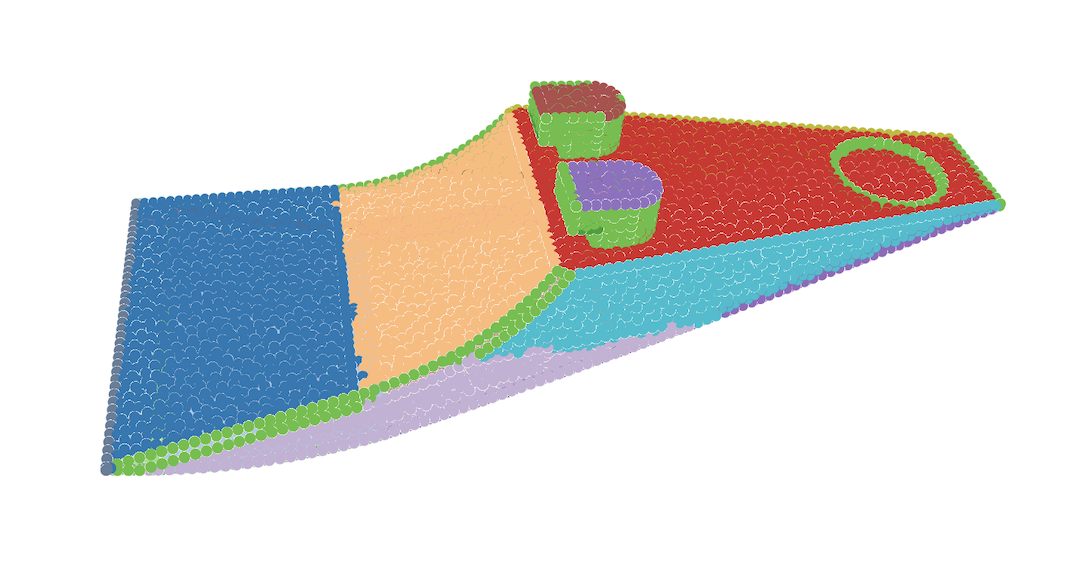}} & 
\TT{\includegraphics[height=0.11\textwidth]      {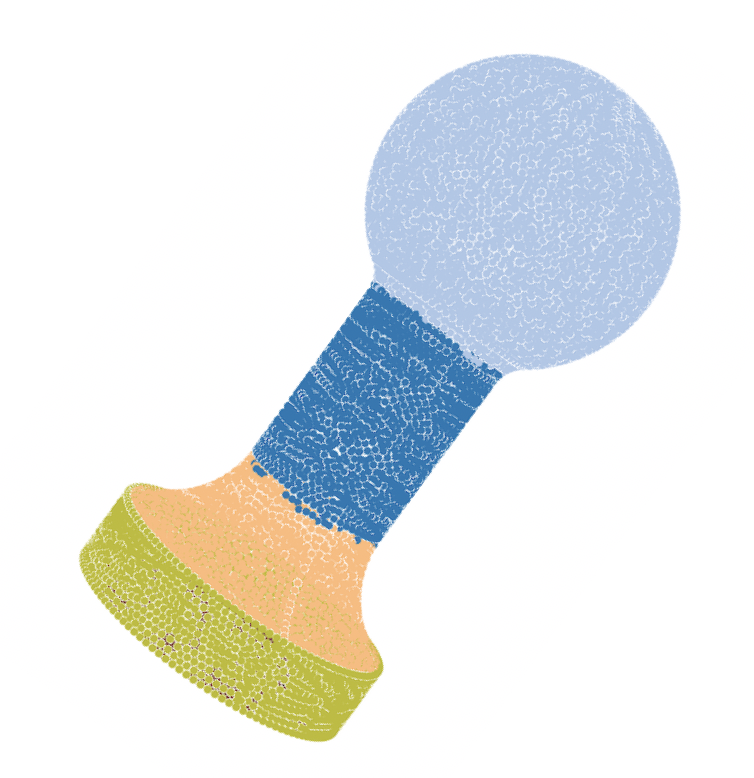}} & 
\TT{\includegraphics[width=\imw]    {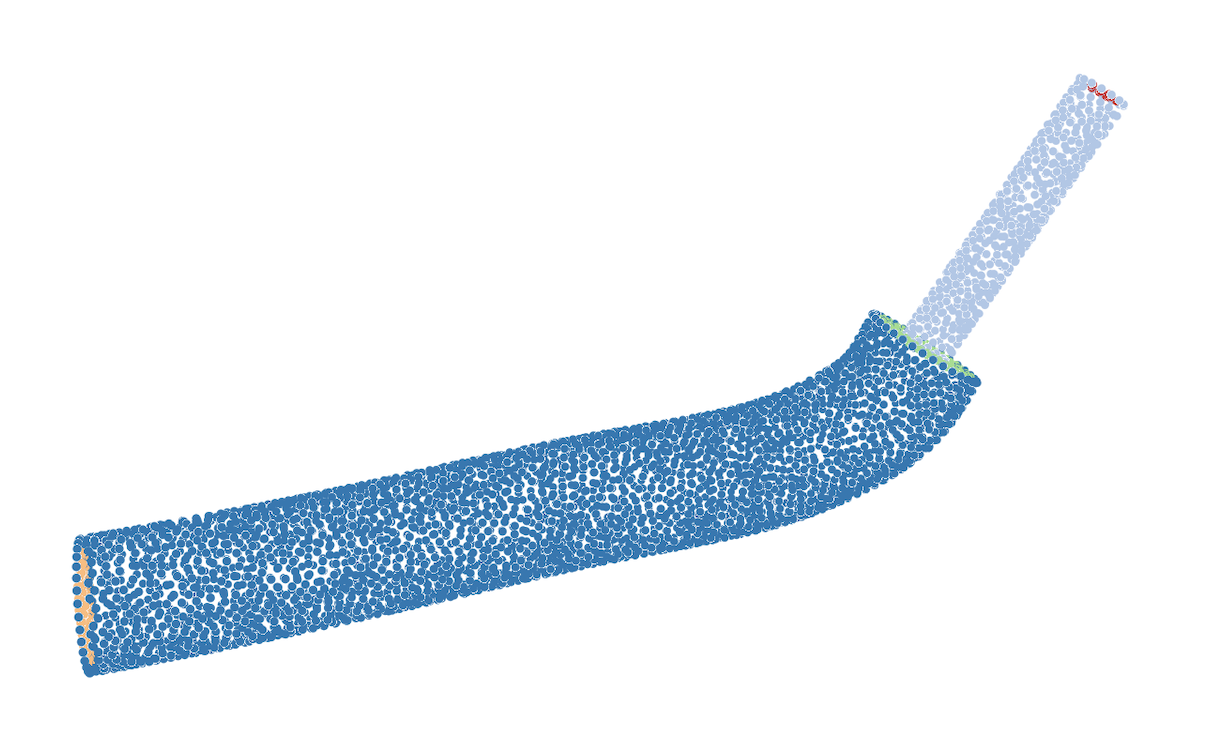}} & 
\TT{\includegraphics[width=0.125\textwidth]    {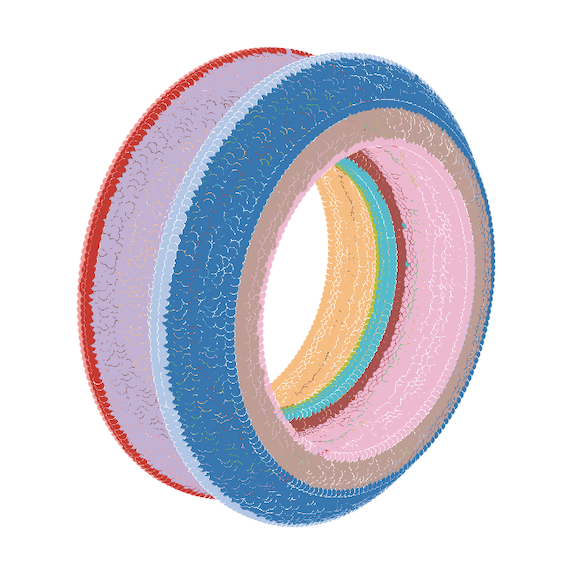}} & 
\TT{\includegraphics[width=0.13\textwidth]      {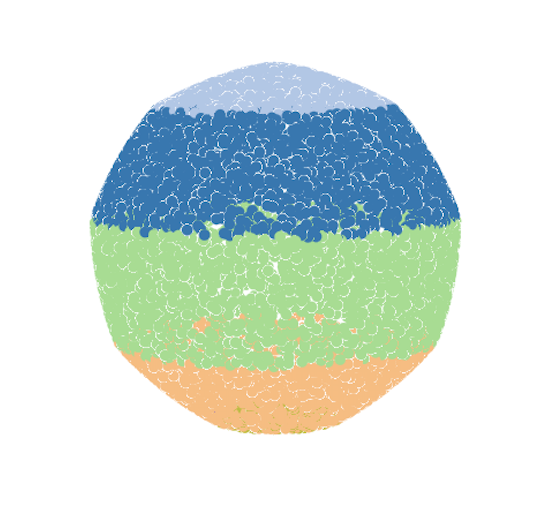}} &
\TT{\includegraphics[width=0.13\textwidth]      {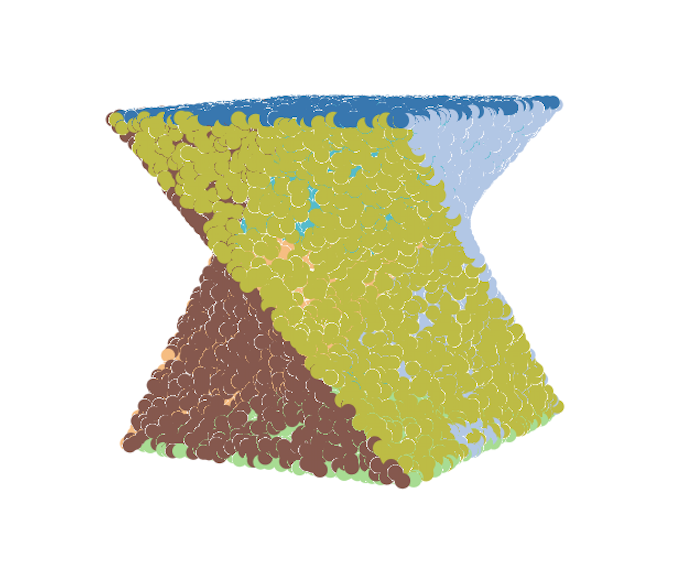}}\\

\end{tabular}
\caption{Primitive segmentation results with different methods. From top to down, we show the results of ground truth, SPFN~\cite{li2019supervised}, ParseNet~\cite{SharmaLMKCM20}, and Our approach.}
\label{Figure:Comparison}
\vspace{-0.15in}
\end{figure*}

\subsection{Network Training}
\label{Subsec:Network:Training}

The network training of HPNet consists of two stages. The prediction module is trained in the first stage, while the hyperparameters of HPNet is learned in the second stage.

\subsubsection{Training of the Prediction Module}

We train the prediction module using a training dataset, in which each point cloud has ground-truth primitive segmentation and associated primitive parameters. In the following, we focus on defining the loss for one point cloud $\set{P}$. The total loss consists of three terms:
\begin{equation}
\set{L}(\set{P}) = \set{L}_{\emb}(\set{P}) + \alpha \set{L}_{\type}(\set{P}) + \beta \set{L}_{\param}(\set{P}),
\end{equation}
where $\set{L}_{\emb}(\set{P})$ is the embedding loss that trains the semantic descriptor; $\set{L}_{\type}(\set{P})$ is the type loss that trains the type vector; $\set{L}_{\param}(\set{P})$ is the parameter loss that trains the primitive parameters. We set $\alpha = 1.0$ and $\beta = 0.1$ in our experiments. Network training employs ADAM~\cite{KingmaB14}. In the following, we define each loss term. 

\vspace{-5pt}
\paragraph{Embedding loss.} Similar to ~\cite{debrabandere2017semantic,yu2019single}, the embedding loss seeks to pull the semantic descriptors close to each other in the same primitive patch and push semantic descriptors of different primitive patches far from each other. Specifically, the loss consists of two terms: $\set{L}_{\pull}$ and $\set{L}_{\push}$. Denote $\set{P}_{k}^{\gt}$ as the ground-truth primitives. $\set{L}_{\pull}$ pulls each descriptor to the mean of the descriptors of the underlying primitive: 
\begin{equation}
\set{L}_{\pull} = \frac{1}{K}\sum^K_{k=1}\frac{1}{|\set{P}_k^{\gt}|}\sum_{p_i\in \set{P}_k^{\gt}}\max\left({\scriptstyle \left\|\bs{d}_i - \bs{d}_{\set{P}_k^{\gt}}\right\| - \delta_1, 0 } \right),
\label{Eq:L:pull}
\end{equation}
where $\bs{d}_{\set{P}_k^{\gt}} = \sum_{p_i\in \set{P}_k^{\gt}}\bs{d}_i/|\set{P}_k^{\gt}|$. $L_{\push}$ pushes the embedding centers away from each other:
\begin{equation}
\set{L}_{\push} = \frac{1}{K(K-1)}\sum_{k < k'}\max\left({\scriptstyle \delta_2 - \left\|\bs{d}_{\set{P}_k^{\gt}} - \bs{d}_{\set{P}_{k'}^{\gt}}\right\|, 0} \right).
\label{Eq:L:push}
\end{equation}
Combing (\ref{Eq:L:pull}) and (\ref{Eq:L:push}), we define the embedding loss as
$$
\set{L}_{\emb} = \lambda \set{L}_{\pull} + \nu \set{L}_{\push}.
$$
In our experiments, $\lambda=1$, $\nu=1$, $\delta_v = 0.5$, and $\delta_d = 1.5$.

\vspace{-10pt}
\paragraph{Type loss.} We employ the cross-entropy loss $H_{ce}$ to define the type loss $$
\set{L}_{\type} = \frac{1}{n}\sum_{i = 1}^n H_{ce}(\bs{t}_i, \bs{t}_i^{\gt}),
$$
where $\bs{t}_i^{\gt}$ is the ground-truth of $\bs{t}_i$.

\vspace{-5pt}
\paragraph{Parameter loss.} The parameter loss employs the standard L2 loss on $\mathbb{R}^{22}$ between the predicted shape parameter $\bs{s}_i$ and the underlying ground-truth $\bs{s}_i^{\gt}$:
$$
\set{L}_{\param} = \frac{1}{n}\sum_{i = 1}^n \|\bs{s}_i - \bs{s}_i^{\gt}\|^2.
$$

\subsubsection{Learning of the HyperParameters}

The second stage optimizes the hyperparameters of HPNet, including those used in defining the spectral modules and those used in defining the entropy terms (Equation~\ref{Eq:Entropy:Fl}). Since the total number of parameters is small, we use the established finite-difference approach for hyperparameter optimization from Song et al.~\cite{SongSH20}. This is done by optimizing the embedding loss on a validation dataset. Given the current hyperparameters, we compute the numerical gradient by sampling neighboring hyperparameter configurations. The best-fitting linear function gives the numerical gradient. We then apply a backtracking line search to determine the step-size. This gradient descent procedure terminates when the step size is smaller than $10^{-3}$, which typically occurs within 10-30 iterations.


\section{Experimental Results}
\label{Section:Experimental:Results}


\subsection{Experimental Setup}
\label{Section:Experimental:Setup}
\paragraph{Datasets.}
We show experimental evaluation on two popular primitive segmentation datasets, i.e., ANSI Mechanical Component Dataset~\cite{li2019supervised} and ABCParts Dataset~\cite{SharmaLMKCM20}. ANSI mainly contains diverse mechanical components provided by TraceParts. Most of the objects in this dataset are composed by four basic primitives(plane, sphere, cylinder, and cone). We have 13k/3k/3k models on train/test/validation sets respectively. Each model contains 8192 points. ABCParts is derived from the ABC dataset~\cite{koch2019abc}, which provides a large source of 3D CAD models. In ABCParts, the objects are more complicated than those in ANSI, and each of them contains at least one B-spline surface patch. We have 24k/4k/4k models on train/test/validation sets on ABCParts and each model contains 10000 points. Please refer to supplementary materials for more details.
\vspace{-5pt}

\begin{figure}
\centering
\includegraphics[width=\columnwidth]{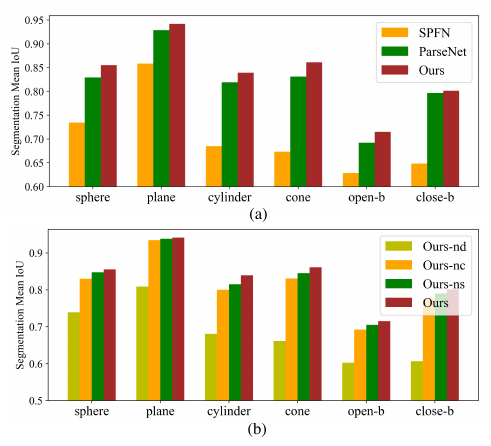}
\caption{Mean IoU of segmentation results on different primitive types. Here, open-b and close-b represent open and closed B-spline patches. (a): comparison between HPNet and baseline methods. (b): comparison between different components of HPNet.}
\label{fig:primitive_type}
\end{figure}

\paragraph{Evaluation metrics.}
We follow the standard metrics proposed by Li \etal~\cite{li2019supervised} for quantitative evaluation.
\begin{itemize}[leftmargin=*,topsep=3pt]
\setlength\itemsep{1pt}
\item \textit{Seg-IoU:} Denoting $K$ as the number of ground-truth patches, this metric evaluates the segmentation mean IoU score: $\frac{1}{K}\sum^{K}_{k=1}\text{IoU}(\mathbf{W}_{:,k}, \hat{\mathbf{W}}_{:,k})$,
where $\mathbf{W}\in \{0,1\}^{n\times K}$ is the predicted segmentation membership matrix; $\hat{\mathbf{W}}\in \{0,1\}^{n\times K}$ is the ground truth.
\item \textit{Type-IoU:} $\frac{1}{K}\sum^K_{k=1}\mathcal{I}[t_k=\hat{t}_k]$, where $t_k$ is the predicted primitive type for the $k$-th segment and $\hat{t}_k$ is the ground truth. $\mathcal{I}$ is an indicator function.
\item \textit{Res-Error:} $\sum^K_{k=1}\frac{1}{m_k}\sum_{\hat{\textbf{s}}_k \in\hat{\set{P}}_k}d_{\set{P}_k}(\hat{\textbf{s}}_k)$, where $\set{P}_k$ is the predicted primitive path, $M_k$ is the number of sampled points $\hat{\textbf{s}}_k$ from ground truth primitive patch $\hat{\set{P}}_k$, $d_{\set{P}_k}(\hat{\textbf{s}}_k)$ is the distance between $\hat{\textbf{s}}_k$ and $\set{P}_k$.
\item \textit{P-coverage:} $\frac{1}{n}\sum^n_{i=1}\delta(\text{min}^K_{k=1}d_{\set{P}_k}(\bs{p}_i) < \epsilon)$, where $\epsilon = 0.01$.
\end{itemize}

\begin{table}

  \centering
  \begin{tabular}{l|c|c|c|c}
  \toprule
  &  \multicolumn{2}{c|}{ANSI} & \multicolumn{2}{c}{ABCParts}\\
  \hline
  &  Seg-IoU & Type-IoU & Seg-IoU & Type-IoU \\
  \hline
  Ours-nd   & 80.10 & 92.45 & 70.33 & 78.87 \\
  Ours-nc  & 92.92 & 98.53 & 83.80 & 89.41 \\
  Ours-ns  & 93.52 & 98.73 & 84.10 & 90.01 \\
  Ours-nw  & 94.07 & \textbf{98.90} & 84.78 & 90.56 \\
  Ours     & \textbf{94.15} & \textbf{98.90} & \textbf{85.24} & \textbf{91.04} \\
  \bottomrule
  \end{tabular}
{%
 \caption{Evaluation on different combinations of our approach. `Ours-nd' denotes dropping the output of the descriptor module.
 `Ours-nc' denotes dropping the output of the consistency module.
 `Ours-ns' denotes dropping the output of the smoothness module.  `Ours-nw' means our approach without weight learning. `Ours' corresponds to our full algorithm.}
 \label{tab:ablation}
}
\end{table}

\subsection{Analysis of Results}
\label{Section:Result:Analysis}

Table~\ref{tab:baseline-comparison} and Figure~\ref{Figure:Comparison} present quantitative and qualitative results of HPNet. We can see that HPNet produces results that are close to the underlying ground-truth. Thanks to the sharp-edge module, the segment boundaries are smooth. Moreover, HPNet can even rectify small over-segmented patches in the training data. Quantitatively, HPNet offers state-of-the-art results under all error metrics on these two datasets.

\paragraph{Baseline comparison.}
Our experimental study considers four baseline approaches. These include two state-of-the-art non-deep learning methods: nearest neighbor (NN)~\cite{Sharma_2018_CVPR} and Efficient RANSAC~\cite{schnabel2007efficient}, and two state-of-the-art deep learning methods: Supervised Primitive Fitting (SPFN)~\cite{li2019supervised} and ParseNet~\cite{SharmaLMKCM20}. Note that to build a fair comparison, we replace the network backbones in SPFN and ParseNet to keep the same as HPNet.

Quantitatively (see Table~\ref{tab:baseline-comparison}), HPNet leads to considerable performance gains from all baseline approaches. Specifically, on ANSI, HPNet leads to salient 12.89\% and 6.30\% improvements in Seg-IOU under the point and point+normal settings, respectively. The improvements under other metrics are also considerable. For example, the Res-Error numbers decrease from 0.013 to 0.011 and from 0.010 to 0.008 under the point and point+normal settings, respectively. The improvements on ABCParts are also considerable, the improvements on Seg-IOU are 4.15\% and 3.77\% under the point and point+normal settings, respectively. HPNet also exhibits consider improvements under other metrics.


Qualitatively (see Figure~\ref{Figure:Comparison}), HPNet leads to better results in the sense that it provides accurate segmentation for both large and small primitive patches. The segmentation boundaries are also smooth. In contrast, baseline approaches ParseNet and SPEN possesses non-smooth boundaries and the issues of over-segmentation and under-segmentation are noticeable.

The relative improvements on ANSI are bigger than those on ABCParts. As illustrated in Figure~\ref{fig:primitive_type}(a), we can understand this behavior from the fact that predictions of shape parameters of B-spline patches are less accurate than the other four primitive types.





\begin{figure}
\centering
\footnotesize
\def\imh{0.3\columewidth}
\def\imw{0.15\textwidth}
\newcommand{\TT}[1]{\raisebox{-0.5\height}{#1}}
\setlength{\tabcolsep}{1pt}
\begin{tabular}{ccccccc}

\TT{\includegraphics[width=\imw]{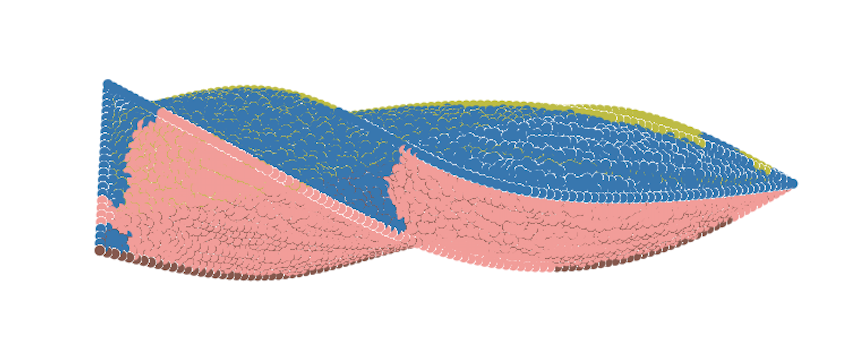}} & 
\TT{\includegraphics[width=\imw]{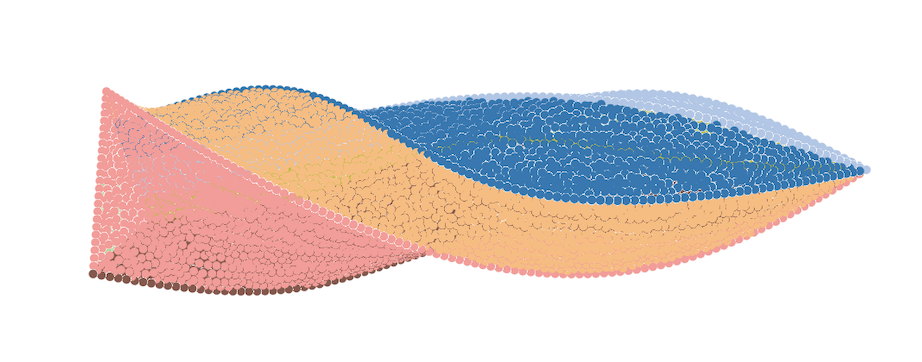}} & 
\TT{\includegraphics[width=\imw]{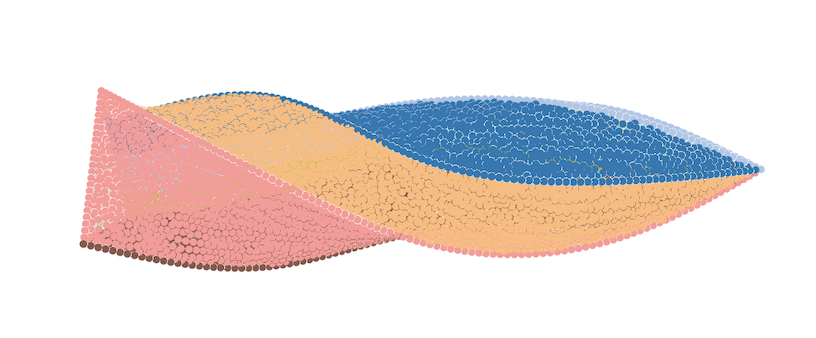}} \\
[+5pt]
\TT{\includegraphics[width=0.10\textwidth]{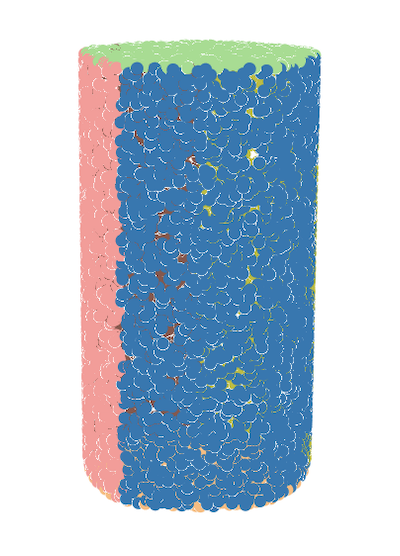}} & \TT{\includegraphics[width=0.11\textwidth]{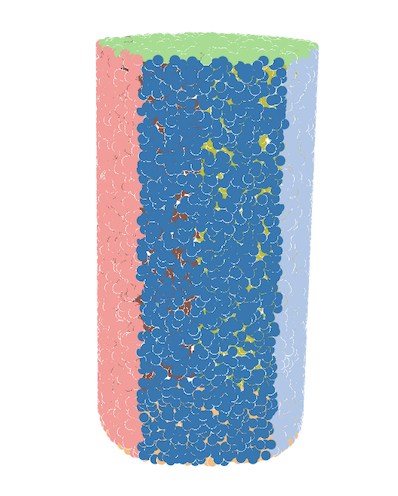}} & 
\TT{\includegraphics[width=0.12\textwidth]{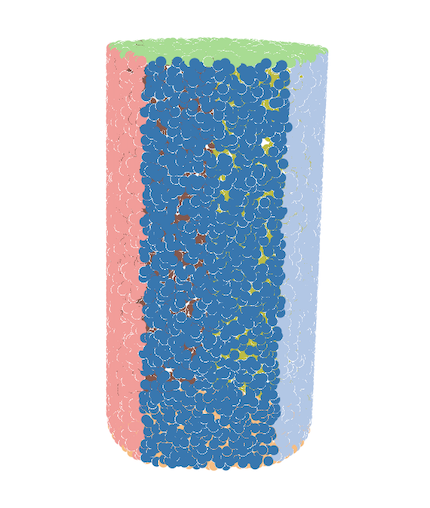}} \\
[+5pt]
\TT{\includegraphics[width=0.1\textwidth]{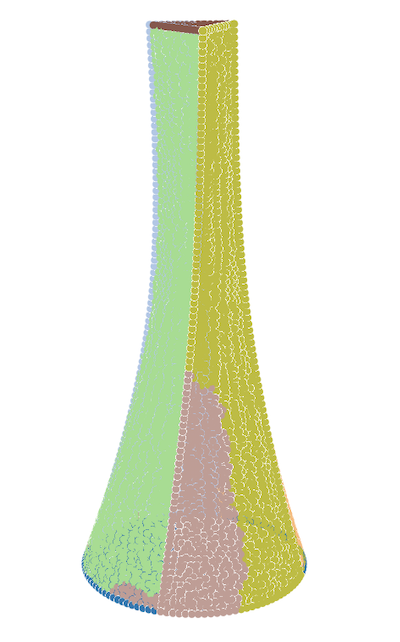}} & 
\TT{\includegraphics[width=0.11\textwidth]{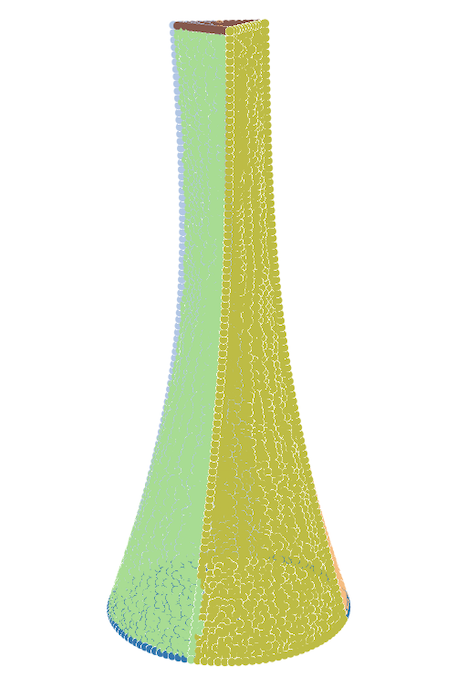}} & 
\TT{\includegraphics[width=0.11\textwidth]{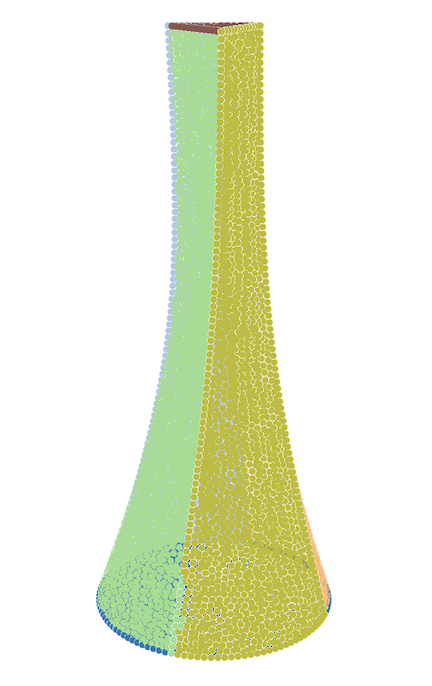}} \\

Ours-ns & Ours & G.T \\
\end{tabular}
\caption{Examples on comparison between with and without sharp edge descriptor. Here, `Ours-ns' represents our model without combining sharp edge descriptor. We notice that adding sharp edge descriptor helps model to capture boundary better.}
\label{fig:boundary}
\vspace{-0.15in}
\end{figure}

\subsection{Ablation Study}
\label{Section:Ablation:Study}

We proceed to study the impacts of different components of HPNet. Table~\ref{tab:ablation} provides the overall Seg-IoU and Type-IoU scores after dropping each component of HPNet. Figure~\ref{fig:primitive_type}(b) shows Seg-IoU scores for different types of primitives.

\paragraph{No descriptor module.} Table~\ref{tab:ablation} shows that the descriptor module is critical for HPNet. Without this module, the Seg-IoU scores drop by 14.9\% and 17.5\% on ANSI and ABCParts, respectively. The Type-IoU scores drop by 6.5\% and 13.4\% on ANSI and ABCParts, respectively. Figure~\ref{fig:primitive_type}(b) shows that except for plane and sphere, the relative performance drops are glaring across other primitive types. This is expected as predicting accurate shape parameters requires global knowledge of the underlying primitive, which can be more difficult than predicting a semantic descriptor. The more significant drop on ABCParts than ANSI can be explained as the fact that ABCParts contain fewer primitives of types sphere and plane.

\vspace{-5pt}
\paragraph{No smoothness module.} As illustrated in Table~\ref{tab:ablation}, dropping the smoothness module leads to 0.67\% and 1.34\% decrements in the Seg-IoU scores and 0.17\% and 1.13\% decrements in the Type-IoU scores on ANSI and ABCParts, respectively. Figure~\ref{fig:primitive_type}(b) shows that the smoothness module contributes most to primitive types of cone, cylinder, and b-spline patches, which are likely to have sharp edges with other primitives. This also explains the smoothness module is slightly more effective on ABCParts than ANSI.

Figure~\ref{fig:boundary} illustrates the effects of the smoothness module. Note that the smoothness module promotes smooth primitive boundaries. It also utilizes the sharp edges to merge and split primitives that are not possible when dropping this module.

\paragraph{No consistency module.} As illustrated in Table~\ref{tab:ablation}, dropping the consistency module leads to 1.31\% and 1.69\% decrements in the Seg-IoU scores and 0.37\% and 2.18\% decrements in the Type-IoU scores on ANSI and ABCParts, respectively. Figure~\ref{fig:primitive_type}(b) shows that the consistency module contributes most to primitive types of \sphere, \cone, and \cylinder patches. One reason is that predictions of shape parameters tend to more accurate on these patches. On the other hand, the accuracy for \plane is already very high, leaving a small margin for improvements.

\begin{figure}
\centering
\includegraphics[width=0.47\textwidth]{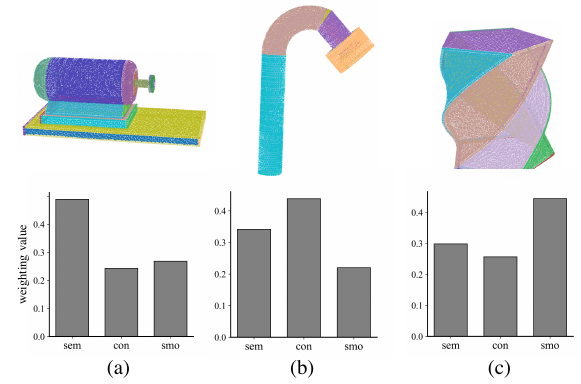}
\caption{Illustration on the learned weights for three typical shapes. For simplicity, we use the L2 norm of the weighting vector of each type of spectral descriptors to represent each spectral embedding space. `sem' denotes semantic descriptor. `con' denotes geometric consistency descriptor. `smo' denotes smoothness descriptor.}
\label{Figure:Weights}
\end{figure}

\paragraph{No weight learning.} Finally, we study the importance of learning combination weights. Table~\ref{tab:ablation} shows that using fixed combination weights leads to 0.08\% and 0.46\% decrements in the Seg-IoU scores and 0.00\% and 0.48\% decrements in the Type-IoU scores on ANSI and ABCParts, respectively. These statistics show that using adaptive weights is effective. Moreover, this strategy has larger impact on ABCParts than ANSI because the diversity of the primitives from ABCParts is larger than that from ANSI.

As shown in Figure~\ref{Figure:Weights}, the learned weights are adaptive for different models. When the model contains small and complex primitives (shape(a)), the semantic descriptor is more useful. If the shape parameter and type predictions are accurate (shape(b)), the consistency descriptor places a more important role. Finally, when sharp edges are prominent (shape(c)), the smoothness descriptor becomes critical.

\section{Conclusions and Limitations}
\label{Section:Conclusions}

In this paper, we have presented HPNet which combines multiple segmentation cues for primitive shape segmentation. Experimental results show that HPNet leads to considerable performance gains compared to previous approaches that leverage a single segmentation cue. Moreover, making the combination weights adaptive to the input models leads to additional performance improvement. Our ablation study further justifies different components of HPNet.

One limitation of HPNet is that it does not utilize symmetric relations among geometric primitives (c.f~\cite{Li:2011:GF}). In the future, we plan to study novel graph neural networks to detect and enforce structural relations among primitives. Detecting such relations is essential for many models such as architectures.

\noindent \textbf{Acknowledgements.} Qixing Huang would like to acknowledge the support from NSF Career IIS-2047677 and NSF HDR TRIPODS-1934932. Etienne Vouga would like to acknowledge the support from NSF IIS-1910274, Side Effects Software Inc., and Adobe Inc.

{\small
\bibliographystyle{ieee_fullname}
\bibliography{paper}
}



\end{document}